\pdfoutput=1
\documentclass[11pt]{article}

\usepackage[preprint]{acl}

\usepackage{times}
\usepackage{latexsym}

\usepackage[T1]{fontenc}
\usepackage[utf8]{inputenc}

\usepackage{microtype}

\usepackage{inconsolata}

\usepackage{graphicx}

\usepackage{microtype}
\usepackage{hyperref}
\usepackage{url}
\definecolor{darkblue}{rgb}{0, 0, 0.5}
\hypersetup{colorlinks=true, citecolor=darkblue, linkcolor=darkblue, urlcolor=darkblue}
\usepackage{dirtytalk}
\usepackage{graphicx}
\usepackage{amsmath}
\usepackage{algorithm}
\usepackage{algpseudocode}
\usepackage{subfigure}
\usepackage{booktabs}
\usepackage{subcaption}
\usepackage{wrapfig}

\title{On the Impact of Noise in Differentially Private Text Rewriting}

\author{Stephen Meisenbacher, Maulik Chevli,\and Florian Matthes \\
Technical University of Munich\\
School of Computation, Information and Technology \\
Department of Computer Science\\
Garching, Germany\\
\texttt{\{stephen.meisenbacher,maulikk.chevli,matthes\}@tum.de} \\
}

\begin{document}
\maketitle
\begin{abstract}
The field of text privatization often leverages the notion of \textit{Differential Privacy} (DP) to provide formal guarantees in the rewriting or obfuscation of sensitive textual data. A common and nearly ubiquitous form of DP application necessitates the addition of calibrated noise to vector representations of text, either at the data- or model-level, which is governed by the privacy parameter $\varepsilon$. However, noise addition almost undoubtedly leads to considerable utility loss, thereby highlighting one major drawback of DP in NLP. In this work, we introduce a new sentence infilling privatization technique, and we use this method to explore the effect of noise in DP text rewriting. We empirically demonstrate that non-DP privatization techniques excel in utility preservation and can find an acceptable empirical privacy-utility trade-off, yet cannot outperform DP methods in empirical privacy protections. Our results highlight the significant impact of noise in current DP rewriting mechanisms, leading to a discussion of the merits and challenges of DP in NLP, as well as the opportunities that non-DP methods present.
\end{abstract}

\section{Introduction}
The field of privacy-preserving Natural Language Processing has gained traction in the research community in response to a growing awareness of the need for data privacy in today's AI landscape, particularly with Large Language Models (LLMs) \cite{9152761,yan2024protecting}. Vulnerabilities of such models have been brought to light by several recent works, mainly in the form of adversarial attacks \cite{9152761, 274574, nasr2023scalable}. Beyond the inherent vulnerabilities of the models themselves, other privacy risks emerge as much of the work for NLP applications has moved off local devices to the cloud-based AI service providers (AI-as-a-Service), made most apparent by the boom in usage of ChatGPT and its competitors \cite{10198233, wu2023unveiling}.

\begin{figure*}[htbp]
    \centering
    \includegraphics[scale=0.5]{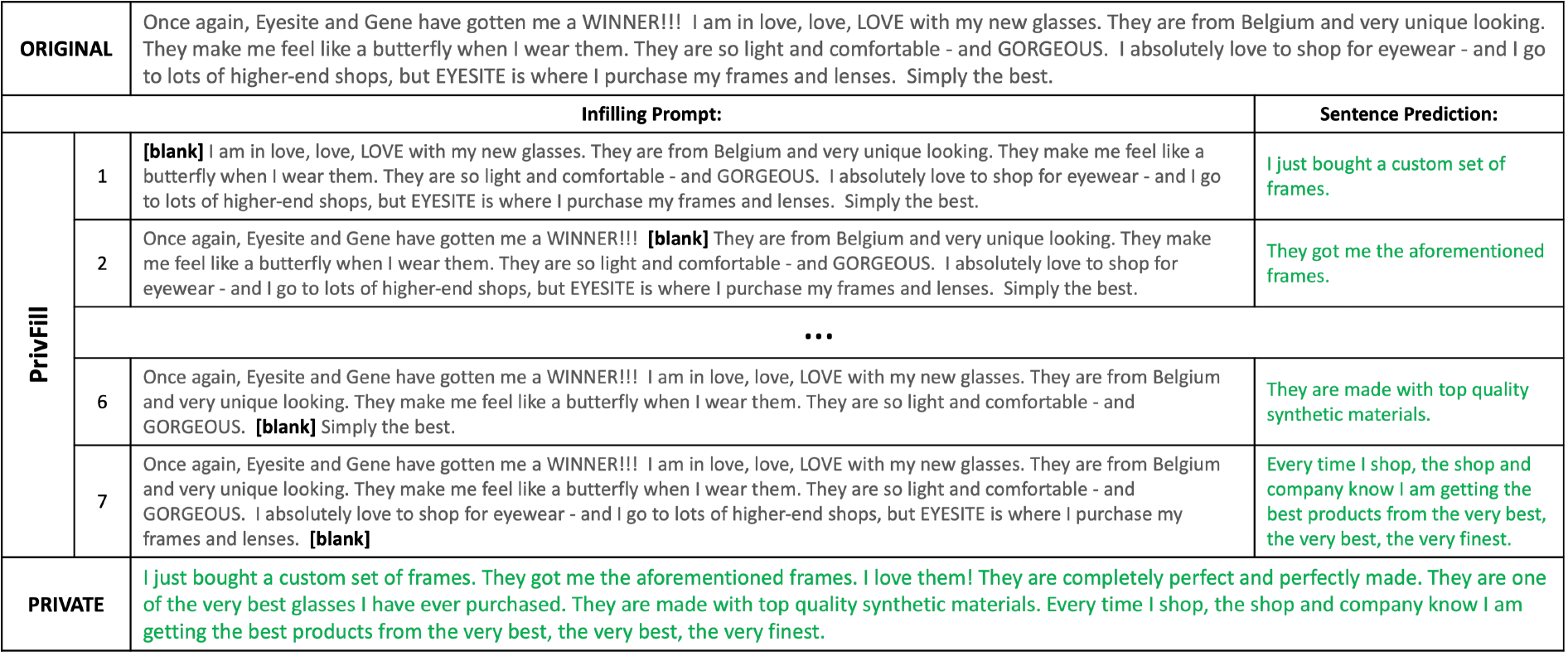}
    \vspace{-15pt}
    \caption{\textsc{PrivFill}: a sentence infilling text rewriting method that leverages generative LMs to create privatized versions of input documents. For each sentence in an input document, the sentence infilling model is tasked with finding a suitable replacement, and these replacement predictions are concatenated to form a privatized rewritten document. (Pictured: example from the Yelp dataset using \textsc{PrivFill} with \textsc{Flan-T5-base}.)}
    \label{fig:privfill}
\end{figure*}

In response to these privacy concerns, numerous solutions have emerged, including the promising and often-utilized privacy-preserving technique of Differential Privacy (DP) \cite{dwork2006differential, desfontaines2020sok}. In essence, DP allows for a quantifiable privacy guarantee, governed by the privacy budget ($\varepsilon$), providing bounded privacy protections for individuals in a dataset. While the study of DP in unstructured domains such as text is less studied and introduces several challenges \cite{klymenko-etal-2022-differential}, numerous technical solutions have been proposed \cite{hu-etal-2024-differentially}, from model-level DP to embedding perturbation methods.

Many DP NLP methods model the integration of \textit{local DP} as a \textit{text rewriting} task \cite{igamberdiev-etal-2022-dp}, in which sensitive input texts are rewritten locally to produce \say{privatized} or \say{perturbed} output texts before release to third parties. Such a task has been realized in a variety of settings, such as word embedding perturbations \cite{10.1145/3336191.3371856}, latent representation perturbations in encoder-decoder models \cite{igamberdiev-habernal-2023-dp}, or DP prompting of LLMs \cite{utpala-etal-2023-locally}. Regardless of where DP is applied, the overarching goal is to produce rewritten texts that are semantically similar to the originals, yet obfuscated enough to mask the direct and indirect information leakage in the originals, thus achieving privacy \cite{klymenko-etal-2022-differential, mattern-etal-2022-limits}.

While DP in the NLP context offers quantifiable privacy protections, it comes with several challenges and caveats \cite{meisenbacher-matthes-2024-thinking}. Among these is clearly reasoning about the syntactic unit protected by DP \cite{klymenko-etal-2022-differential, vu-etal-2024-granularity}, ensuring syntactic and grammatical correctness while privatizing \cite{mattern-etal-2022-limits}, or high-dimensional spaces leading to significantly higher $\varepsilon$ values for rewritten outputs to be semantically meaningful \cite{igamberdiev-habernal-2023-dp}. Notably, the necessity to add considerable noise to text representations to fulfill the DP guarantee presents a challenge to achieving practically usable privatization mechanisms. This challenge is exacerbated when privatizing longer text documents \cite{utpala-etal-2023-locally}.

In this work, we investigate these challenges by measuring the effect of noise in DP text rewriting. With this, we pose the following research question:
\begin{quote}
    \textit{Can language models be leveraged without noise addition for the task of private text rewriting, and what are the empirical utility and privacy results as compared to DP methods with noise?}
\end{quote}

Language models have been leveraged for text privation to achieve DP text rewriting in SOTA methods like DP-Prompt \cite{utpala-etal-2023-locally}. These methods, however, utilize LMs to generate privatized texts as a continuation of the original sensitive texts, for example via zero-shot text paraphrasing. This can lead to privatized texts that are considerably shorter than the originals, thereby losing valuable context. To address this, we introduce a new text privatization method based on sentence infilling that allows for greater control and higher preserved context in text rewriting over using a standard prompting technique for text completion. 

We introduce a non-DP and DP version of our method, \textsc{PrivFill}, allowing for direct comparisons to previously proposed DP mechanisms. We argue that the exhibited ability of our non-DP \textsc{PrivFill} variant to act as a usable privatization mechanism makes a compelling case for adoption as opposed to utility-degrading DP methods that must be carefully tuned via $\varepsilon$ for desired results. The empirical results, however, showcase the strong privacy-preserving capabilities of mechanisms \textit{with} DP guarantees, providing further empirical evidence of the fine balance between privacy and utility.

The contributions of our work are as follows:
\begin{enumerate}
    \itemsep -0.3em
    \item We introduce a text privatization mechanism (\textsc{PrivFill}), which leverages language model sentence infilling for private text rewriting.
    \item We evaluate \textsc{PrivFill}, both with and without DP guarantees, in a series of utility and privacy experiments, in comparison to state-of-the-art DP rewriting mechanisms.
    \item We showcase the merits of text privatization without noise addition, and we critically discuss the advantages of such an approach in juxtaposition to DP mechanisms.
    \item We open-source \textsc{PrivFill}, which can be found at: \url{https://github.com/sjmeis/PrivFill}
\end{enumerate}

\section{Related Work}
\subsection{Differential Privacy in NLP}
Differential Privacy in NLP often leverages the notion of \textit{local} DP \cite{feyisetan_balle_2020,klymenko-etal-2022-differential,hu-etal-2024-differentially,meisenbacher2024comparative}, in which text is privatized via some local mechanism, i.e., on the user's device. This is especially applicable for text rewriting, which does not necessitate user texts to be first collected at some central location as would be required for techniques like DP-SGD \cite{Abadi2016DeepLW,Kerrigan2020DifferentiallyPL,ponomareva-etal-2022-training}.

Text rewriting approaches that integrate DP primarily fall under the category of \textit{embedding perturbation} methods \cite{hu-etal-2024-differentially}. Earlier methods first targeted the \textit{word-level} as the unit of protection. \citet{fernandes2019generalised} and \citet{feyisetan_balle_2020} proposed the usage of \textit{metric} DP for word-level application, leading to a series of follow-up works that aimed to strengthen the utility preservation of word-level DP mechanisms \cite{yue,Chen2022ACT,carvalho2023tem,meisenbacher-etal-2024-dp}. As pointed out by \citet{mattern-etal-2022-limits}, word-level approaches lack contextualization and often struggle with grammatical correctness. As such, other DP NLP works operate directly at the sentence- or document-level \cite{Bo2019ERAEDP,10.1145/3485447.3512232,Meehan2022SentencelevelPF,igamberdiev-habernal-2023-dp}. State-of-the-art methods leverage larger language models for DP text rewriting, using the \textit{paraphrasing} task as a proxy \cite{mattern-etal-2022-limits, utpala-etal-2023-locally}. The main limitation of DP mechanisms targeting document-level text representations, however, is the requirement for high privacy budgets (e.g., 1000) to create meaningful outputs, as highlighted by \citet{igamberdiev-habernal-2023-dp}. 
%Such a limitation calls into question the merits of a theoretical privacy guarantee when the guarantee is greatly weakened by a significantly high $\varepsilon$ value (e.g., 1000). 

\subsection{Non-DP Text Privatization}
Beyond DP, stylometric-based methods have been used to replace sensitive words in a text to prevent information leakage of attributes that can be traced to an individual \cite{reddy-knight-2016-obfuscating,Shen2017StyleTF}. Cyclic machine translation is one of the earliest and still prevalent ways to obfuscate text \cite{prabhumoye-etal-2018-style,Xu2019PrivacyAwareTR,Adelani2021PreventingAP}. More recently, style transfer techniques have been used to generate a controlled, obfuscated text using a generative model \cite{Shen2017StyleTF,He2020A,Tokpo2022TextST}. These privatization methods have not explored the usage of (large) language models; accordingly, we propose a method that addresses this gap, specifically as an alternative to DP methods.

\subsection{Text Infilling}
Language Models (LMs), particularly recent LLMs, possess strong generative abilities when presented with a text input and tasked to continue or complete the text. %In the general case, \textit{autoregressive} LMs (e.g., GPT) and encoder-decoder LMs (e.g., BART or T5) generate a continuation of text input token-by-token, where each generation represents a sampling of the most likely next token prediction as gauged by the model's weights. While this is a common use case and is especially effective in \textit{prompting} LLMs, 
The task of \textit{infilling} aims to transform text generation to predict the most likely replacement for some missing span at any point in an input text.
Modern text infilling is performed by fine-tuning a generative LM for the task of missing span (sentence or document) prediction. \citet{donahue-etal-2020-enabling} propose a simple method for training an LM to fill in any number of missing tokens within a text, and they extend this method to replace entire missing spans. \citet{huang-etal-2020-inset} and \citet{shen-etal-2020-blank} propose similar methods.% that allow for control over generation. %More recent works have built upon the foundations of infilling by LMs, adding components such as a separate discriminator or enhanced positional embeddings to guide the generation \cite{li-etal-2022-tip, du-etal-2022-glm}.

%The effectiveness of infilling mechanisms is often evaluated with human annotators, for example by measuring the fluency or coherence of the generated outputs. 
Interestingly, \citet{donahue-etal-2020-enabling} reports that humans have a difficult time discerning which outputs are created by infillers and which are not. Furthermore, results also indicate that humans sometimes \textit{prefer} rewritten versions over the original \cite{huang-etal-2020-inset}. This motivates the use of infillers for text privatization via a rewriting task, for the purpose of creating coherent, yet distinct and semantically plausible replacements for text documents. To the best of our knowledge, this is the first approach to view the task of text infilling in the same light as private text rewriting. In particular, we focus on sentence infilling, finding the balance between computationally heavy word infilling and potentially incoherent document infilling.

\section{A Simple Infilling-based Text Privatization Mechanism}
We introduce \textsc{PrivFill}, a simple text rewriting mechanism that rewrites text documents by leveraging fine-tuned LM infilling models. We outline the process undertaken for infiller training, as well as describe the underlying mechanism of \textsc{PrivFill}.

\begin{figure*}[htbp]
    \centering
    \includegraphics[scale=0.45]{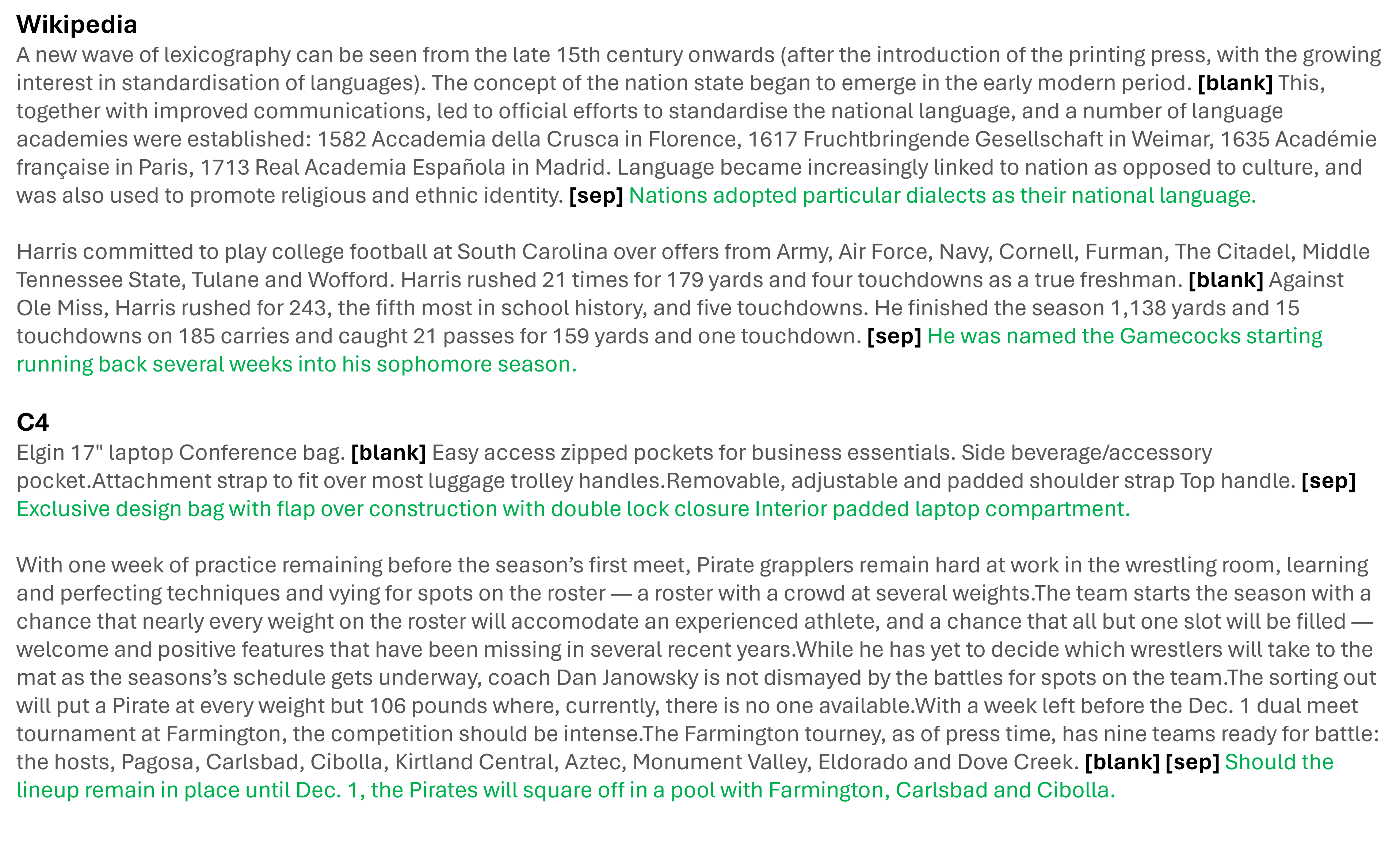}
    \vspace{-10pt}
    \caption{Samples from our infilling train dataset (target infilling text in green, \textbf{[sep]} token inserted for readability).}
    \label{fig:train}
\end{figure*}

\subsection{Training a Sentence Infiller}
To train our base sentence infilling models for \textsc{PrivFill}, we follow the approach proposed by \citet{donahue-etal-2020-enabling}, slightly modified for our objectives. To create training samples for sentence infilling, we employ two datasets: English Wikipedia and Common Crawl. The exact dataset preparation details for each dataset are described next.

\paragraph{Wikipedia.}
We select the train split of the \texttt{20220301.en} version of the English Wikipedia dump \cite{wikidump}. Next, we randomly sample one million articles, and from each, we randomly select one sentence to mask out. Due to the potentially large size of articles, we only keep the preceding two and ensuing two context sentences, if available. Thus, the resulting infilling dataset contains one million samples where the missing sentence is \say{centered} in context.

\paragraph{Common Crawl (C4).}
We employ the \textit{Colossal Clean Crawled Corpus} (C4) made available by \citet{JMLR:v21:20-074}, using the \texttt{en} (cleaned) split. We randomly sample one million rows, and within each sampled text, we randomly select one sentence to mask. As opposed to Wikipedia, we preserve the entire text, due to the smaller average text length.

\paragraph{Final Dataset.}
The final infilling train dataset is created by merging the two abovementioned prepared datasets. As done by \citet{donahue-etal-2020-enabling}, we utilize the \texttt{[blank]} token to represent the masked-out sentences. We do not use the \texttt{[answer]} token to demarcate the target text spans, as we only mask one sentence per training instance. Examples originating from both training corpora can be found in Figure \ref{fig:train}. 

\subsubsection{Model Training}
We choose three base model checkpoints for fine-tuning on our infilling dataset: \textsc{bart-large} \cite{DBLP:journals/corr/abs-1910-13461}, \textsc{flan-t5-base}, and \textsc{flan-t5-large} \cite{https://doi.org/10.48550/arxiv.2210.11416}. These models were chosen for two reasons: (1) they are powerful generative models that are still (relatively) small in size, thus not too resource-intensive, and (2) they present an opportunity to compare with selected DP methods, discussed in Section \ref{sec:selected}.

For fine-tuning, we use the HuggingFace Trainer. All models were trained for one epoch with a learning rate of $5\mathrm{e}\text{-}{5}$ on a 99\% training split of our infilling dataset. The ROUGE metric \cite{lin-2004-rouge} was used for evaluation. All other default training parameters were kept. For performance, we limit all models to a max length of 512 tokens. %\footnote{\url{https://huggingface.co/docs/transformers/en/main_classes/trainer}}

\subsection{\textsc{PrivFill}}
We now introduce our \textsc{PrivFill} mechanism, a simple infilling privatization mechanism that rewrites text documents sentence-by-sentence.

\begin{algorithm}[htbp]
\small
\caption{\newline \small \textsc{PrivFill} Text Rewriting Mechanism}
    \label{alg:privfill}
    \begin{algorithmic}
        \Require Infilling Model $\mathcal{M}$, Tokenizer $\mathcal{T}$, input document $d$, max length $max\_len$, max new tokens $mnt$
        \Ensure privatized document $priv$
        \State $\texttt{sentences} \gets sent\_tokenize(d)$
        \State $\texttt{rewritten} \gets [ \ ]$
        \For {$idx \in 1...len(\texttt{sentences})$}
            \State $\texttt{prompt} \gets d.replace(\texttt{sentences}[idx], ``\texttt{[blank]}'')$
            \State $\texttt{inputs} \gets \mathcal{T}.encode(\texttt{prompt}, max\_len)$
            \State $\texttt{output} \gets \mathcal{M}.generate(\texttt{inputs}, mnt, do\_sample)$
            \State $\texttt{decoded} \gets \mathcal{T}.decode(\texttt{output})$
            \State $\texttt{rewritten}.append(\texttt{decoded})$
        \EndFor
        \State $priv = concatenate(\texttt{rewritten})$
        \State \Return $priv$
    \end{algorithmic}
\end{algorithm}

As specified in Algorithm \ref{alg:privfill}, the \textsc{PrivFill} mechanism rewrites a document in sequential order of the sentences from the input document. The privatized output document is thus comprised of the predicted sentences from the infilling model for each of the component sentences. In this way, we can create privatized documents composed of contextualized sentence replacements. Due to the nature of infilling, \textsc{PrivFill} would require documents with at least two sentences for sensible outputs, although this is not a necessary condition.

\begin{algorithm}[ht!]
\small
\caption{\newline \small \textsc{PrivFillDP} Text Rewriting Mechanism}
    \label{alg:privfilldp}
    \begin{algorithmic}
        \Require Infilling Model $\mathcal{M}$, Tokenizer $\mathcal{T}$, input document $d$, max length $max\_len$, max new tokens $mnt$, privacy budget $\varepsilon$, sensitivity $\Delta$
        \Ensure privatized document $priv$
        \State $\texttt{temperature} \gets \frac{2\cdot \Delta}{\varepsilon}$
        \State $\texttt{sentences} \gets sent\_tokenize(d)$
        \State $\texttt{rewritten} \gets [ \ ]$
        \For {$idx \in 1...len(\texttt{sentences})$}
            \State $\texttt{prompt} \gets d.replace(\texttt{sentences}[idx], ``\texttt{[blank]}'')$
            \State $\texttt{inputs} \gets \mathcal{T}.encode(\texttt{prompt}, max\_len)$
            \State $\texttt{logits} \gets \mathcal{M}.encode(\texttt{inputs})$
            \State $\texttt{logits} \gets \texttt{logits} / \texttt{temperature}$
            \State $\texttt{output} \gets \mathcal{M}.generate(\texttt{logits}, mnt, do\_sample)$
            \State $\texttt{decoded} \gets \mathcal{T}.decode(\texttt{output})$
            \State $\texttt{rewritten}.append(\texttt{decoded})$
        \EndFor
        \State $priv = concatenate(\texttt{rewritten})$
        \State \Return $priv$
    \end{algorithmic}
\end{algorithm}

\subsection{\textsc{PrivFillDP}}
We also construct a variant of \textsc{PrivFill} that fulfills DP guarantees, in order to evaluate the effect of DP (i.e., noise). To transform \textsc{PrivFill} into \textsc{PrivFillDP}, we utilize a DP token selection mechanism based on the Exponential Mechanism \cite{mattern-etal-2022-limits, utpala-etal-2023-locally, meisenbacher-etal-2024-dp}. Specifically, we use the fine-tuned \textsc{flan-t5-large} infilling model described previously, and in the generation phase, tokens are sampled in the same way as with \textsc{DP-Prompt} (see Section \ref{sec:prompt}); hence, the theoretical DP guarantees are the same as DP-Prompt. We use the same clipping bounds and $\varepsilon$ values as reported in Section \ref{sec:prompt}, to ensure direct comparability to a SOTA DP mechanism. For further details on the DP token selection mechanism based on temperature sampling, we refer the reader to the original works mentioned above and to Algorithm \ref{alg:privfilldp}, which outlines the pseudocode for \textsc{PrivFillDP}.

\section{Experimental Setup}
We describe our experimental design, specifically the tasks assigned to both utility and privacy experiments. These experiments evaluate the utility- and privacy-preserving capabilities of \textsc{PrivFill} comparatively to previous DP rewriting methods.

\subsection{Selected DP Mechanisms}
\label{sec:selected}
%As points of comparison from the field of private text rewriting with DP, two state-of-the-art methods from the recent literature are chosen to compare with \textsc{PrivFill}.

\paragraph{DP-BART \cite{igamberdiev-habernal-2023-dp}.}
DP-BART introduces a DP mechanism to the latent representation between the encoder and decoder blocks of a \textsc{BART} model \cite{DBLP:journals/corr/abs-1910-13461}. In essence, a noisy encoder representation is fed to the \textsc{BART} decoder, which then outputs a privatized version of the original text. We test the \textbf{DP-BART-CLV} version as proposed by \citet{igamberdiev-habernal-2023-dp}, as this does not require further training. For $\varepsilon$ values, we choose $\varepsilon \in \{500, 1000\}$, as these roughly represent the median and third quartile of evaluated values in the original paper.

\paragraph{DP-Prompt \cite{utpala-etal-2023-locally}.}
\label{sec:prompt}
DP-Prompt leverages zero-shot prompting of LLMs to rewrite texts by \textit{paraphrasing}. In particular, DP is ensured by the utilization of a temperature sampling mechanism that mimics the behavior of the well-known Exponential Mechanism \cite{4389483}. The authors of DP-Prompt test their mechanism on a variety of open- and closed-source LLMs. For our evaluation, we utilize \textsc{Flan-T5-Large} \cite{https://doi.org/10.48550/arxiv.2210.11416} due to it being open-source, not too resource intensive (780M parameters), and directly comparable to our \textsc{PrivFill} base models. 

For DP-Prompt, the calculation and choice of $\varepsilon$ value depends on the clipping range chosen for the model logit values. Following \citet{utpala-etal-2023-locally}, we measure the observed range of logit values for our chosen \textsc{Flan-T5-Large} model, run on 100 randomly selected texts from the C4 corpus, and accordingly choose the clipping range to be $(logit_{min}, logit_{max}) = (-95, 8)$, which is normalized to the range of $[0,1]$\footnote{as done by \citet{mattern-etal-2022-limits}}, thus leading to a sensitivity of $1$. Next, we choose the temperature values $T$ of 1.0 and 1.5, representing the first and third quartile values used in the original implementation\footnote{As only five values were tested by \citet{utpala-etal-2023-locally}, we opt for different quartile values than with DP-BART.}. Under the temperature sampling mechanism, this equates to values of $\varepsilon \in \{1.\overline{3}, 2\}$\footnote{Following the formula $\varepsilon = \frac{2 \cdot \Delta}{T}$, where $\Delta$ represents sensitivity. Here, $\varepsilon$ is \textit{per generated token}.}.

\begin{table*}[htbp]
\resizebox{\textwidth}{!}{
\begin{tabular}{l|ccccc|ccccc|ccccc}
\multicolumn{1}{r|}{Task} & \multicolumn{5}{c|}{arXiv (10 classes)} & \multicolumn{5}{c|}{BBC (5 classes)} & \multicolumn{5}{c}{DocNLI (2 classes)} \\ \hline
\multicolumn{1}{r|}{Baseline} & \multicolumn{5}{c|}{$46.58_{1.0}$ (42)} & \multicolumn{5}{c|}{$98.52_{0.3}$ (24)} & \multicolumn{5}{c}{$86.87_{1.8}$ (48)} \\ \hline
\multicolumn{1}{l|}{Mechanism} & R1 & RL & CS & PPL & F1 $\uparrow$ & R1 & RL & CS & PPL & F1 $\uparrow$ & R1 & RL & CS & PPL & F1 $\uparrow$ \\ \hline
DP-BART ($\varepsilon=500$ / document) & 0.21 & 0.14 & 0.08 & 26 & $29.53_{0.0}$ & 0.25 & 0.14 & 0.17 & 24 & $46.67_{1.3}$ & 0.21 & 0.14 & 0.18 & 29 & $75.10_{2.5}$ \\
DP-BART ($\varepsilon=1000$ / document) & 0.30 & 0.18 & 0.22 & 11 & $30.81_{1.8}$ & 0.36 & 0.17 & 0.45 & 11 & $89.42_{1.0}$ & 0.34 & 0.21 & 0.47 & 12 & $\mathbf{76.48_{0.5}}$ \\
DP-Prompt ($\varepsilon=1.\overline{3}$ / token) & 0.05 & 0.04 & 0.17 & 8773 & $29.53_{0.0}$ & 0.04 & 0.02 & 0.20 & 9895 & $29.31_{1.9}$ & 0.05 & 0.03 & 0.20 & 10762 & $54.34_{3.4}$ \\
DP-Prompt ($\varepsilon=2$ / token) & 0.11 & 0.08 & 0.58 & 612 & $34.78_{1.0}$ & 0.07 & 0.06 & 0.59 & 809 & $88.15_{0.6}$ & 0.11 & 0.09 & 0.48 & 1283 & $60.14_{3.5}$ \\ \hline
\textsc{PrivFillDP} ($\varepsilon=1.\overline{3}$ / token) & 0.13 & 0.06 & 0.20 &  7945 & $31.48_{1.4}$ & 0.18 & 0.07 & 0.35 & 6856 & $89.95_{0.5}$ & 0.12 & 0.06 & 0.25 & 9376 & $69.47_{2.9}$ \\
\textsc{PrivFillDP} ($\varepsilon=2$ / token) & 0.32 & 0.14 & 0.53 & 468 & $34.69_{0.3}$ & 0.39 & 0.14 & 0.65 & 309 & $96.61_{0.4}$ & 0.29 & 0.13 &  0.52 & 496 & $71.66_{0.6}$ \\ \hline
\textsc{PrivFill} (\textsc{bart-large}) & 0.55 & 0.28 & 0.83 & 15 & $40.10_{4.2}$ & 0.54 & 0.25 & 0.82 & 9 & $97.04_{0.4}$ & 0.47 & 0.26 & 0.75 & 15 & $73.71_{1.8}$ \\
\textsc{PrivFill} (\textsc{flan-t5-base}) & 0.47 & 0.21 & 0.76 & 48 & $39.12_{4.2}$ & 0.51 & 0.19 & 0.80 & 31 & $97.99_{0.7}$ & 0.40 & 0.18 & 0.69 & 46 & $74.51_{2.5}$ \\
\textsc{PrivFill} (\textsc{flan-t5-large}) & 0.48 & 0.21 & 0.77 & 45 & $\mathbf{42.18_{2.9}}$ & 0.51 & 0.19 & 0.80 & 27 & $\mathbf{98.20_{0.4}}$ & 0.41 & 0.18 & 0.70 & 40 & $73.41_{0.7}$
\end{tabular}
}
\caption{Utility Results. \textit{R1}, \textit{RL}, \textit{CS}, and \textit{PPL} denote ROUGE-1/-L, cosine similarity, and perplexity, respectively. The highest F1 for each task is \textbf{bolded}. Subscripts denote standard deviations. Baseline PPL is given in parentheses.}
\label{tab:utility}
\end{table*}

\subsection{Utility Experiments}
We employ three datasets to test the utility preservation of our selected mechanisms on document classification tasks. The results of these experiments are found in Table \ref{tab:utility}.

\subsubsection{Datasets and Tasks}
\paragraph{arXiv Abstracts.}
The \textit{arXiv Paper Abstracts} dataset\footnote{\url{https://www.kaggle.com/datasets/spsayakpaul/arxiv-paper-abstracts}} is a collection of nearly 40,000 arXiv papers, specifically the title and abstract. Each paper is classified into one or more arXiv paper categories. We only consider papers belonging to the top 10 most frequent categories, namely \{\textit{cs.AI}, \textit{cs.LG}, \textit{stat.ML}, \textit{cs.CV}, \textit{cs.CL}, \textit{eess.IV}, \textit{cs.RO}, \textit{cs.CR}, \textit{cs.NE}, \textit{cs.GR}\}. Papers with multiple labels are considered only by their top (first) label. This results in a final dataset of 26,109 rows. With this dataset, we create a 10-class single-label classification task.

\paragraph{BBC News.}
The \textit{BBC News} dataset\footnote{\url{http://mlg.ucd.ie/datasets/bbc.html}} is a corpus of 3147 BBC News articles from five categories: \{\textit{business}, \textit{entertainment}, \textit{politics}, \textit{sports}, \textit{tech}\}, creating a 5-class single-label classification task.

\paragraph{DocNLI.}
The \textit{DocNLI} dataset \cite{yin-etal-2021-docnli} was created to introduce a \textit{document}-level entailment prediction task, as opposed to previous approaches focusing on sentences. The dataset consists of (\textit{premise}, \textit{hypothesis}) pairs, each marked as \textit{entailment} or \textit{not entailment}. Since the original dataset is expansive, we take a 1\% random sample, resulting in a 9136-row dataset. The dataset is then used for a two-class binary classification task.

\subsubsection{Evaluation Strategy}
For each of the three datasets, we first achieve baseline (micro) F1 Scores by fine-tuning a \textsc{deberta-v3-base} \cite{he2021deberta} model on the original datasets. Each dataset is separated into a 90/10 split. The model is trained on the train split for one epoch using a learning rate of $5\mathrm{e}\text{-}{5}$ and otherwise default HuggingFace \textsc{Trainer} parameters. Training is repeated three times with a different shuffled version of the train split. All reported metrics represent the trained model's performance on the validation split, averaged over the three evaluations.

Next, each of the privatization approaches is used to rewrite the datasets. These rewritten datasets are then used for the same fine-tuning procedure, and the final evaluation metrics are captured for comparison to the baseline. Thus, the experiments measure the ability of each rewriting approach to preserve the utility of the original data. 

\subsubsection{Additional Metrics}
In addition to F1, we report three additional metrics usually employed in text generation evaluation: ROUGE \cite{lin-2004-rouge} (i.e., ROUGE-1/-L), cosine similarity (CS, inspired by BERTScore \cite{bert-score}, and perplexity (PPL). While not explicit measures of utility preservation, these three metrics are useful in explaining the relation between the original and rewritten texts, e.g., whether they are similar in (n-gram) token overlap and/or semantic similarity. As such, the scores will be useful to the analysis of the results presented later in this section. Note that for the CS score, we use the \textsc{all-MiniLM-L6-v2} model from \textsc{sentence-transformers}. PPL is used to measure the \textit{plausibility} and \textit{naturalness} of the privatized texts \cite{10.1145/3485447.3512232, 10.1145/3664476.3669926}. We report mean PPL using a \textsc{GPT-2} model. \cite{radford2019language}.

\begin{table*}[htbp]
\centering
    %\begin{subtable*}
        \resizebox{\textwidth}{!}{
        \begin{tabular}{l|c|ccccccccc}
 \textbf{Trustpilot (Gender)} & \multicolumn{1}{c|}{Baseline} & \multicolumn{1}{c}{\begin{tabular}[c]{@{}c@{}}DP-BART\\ ($\varepsilon=500$)\end{tabular}} & \multicolumn{1}{c}{\begin{tabular}[c]{@{}c@{}}DP-BART\\ ($\varepsilon=1000$)\end{tabular}} & \multicolumn{1}{c}{\begin{tabular}[c]{@{}c@{}}DP-Prompt\\ ($\varepsilon=1.\overline{3}$)\end{tabular}} & \multicolumn{1}{c}{\begin{tabular}[c]{@{}c@{}}DP-Prompt\\ ($\varepsilon=2$)\end{tabular}} & \multicolumn{1}{c}{\begin{tabular}[c]{@{}c@{}}\textsc{PrivFillDP}\\ ($\varepsilon=1.\overline{3}$)\end{tabular}} & \multicolumn{1}{c}{\begin{tabular}[c]{@{}c@{}}\textsc{PrivFillDP}\\ ($\varepsilon=2$)\end{tabular}} & \multicolumn{1}{c}{\begin{tabular}[c]{@{}c@{}}\textsc{PrivFill}\\ (\textsc{bart-large})\end{tabular}} & \multicolumn{1}{c}{\begin{tabular}[c]{@{}c@{}}\textsc{PrivFill}\\ (\textsc{flan-t5-base})\end{tabular}} & \multicolumn{1}{c}{\begin{tabular}[c]{@{}c@{}}\textsc{PrivFill}\\ (\textsc{flan-t5-large})\end{tabular}} \\ \hline
Utility F1 $\uparrow$ & $99.57_{0.1}$ & $93.59_{0.2}$ & $98.16_{0.0}$ & $92.18_{0.3}$ & $95.13_{0.1}$ & $93.53_{1.2}$ & $96.80_{0.1}$ & $98.63_{0.1}$ & $97.33_{0.2}$ & $98.73_{0.1}$  \\
PP+$\uparrow$ & 289 & -309 & 148 & -450 & -155 & -315 & 12 & 195 & 65 & 205 \\
Privacy F1 (static) $\downarrow$ & 72.46 & 59.61 & 59.47 & 58.56 & 62.16 & 58.12 & 62.26 & 60.16 & 61.00 & 60.73 \\ 
Privacy F1 (adapt.) $\downarrow$ & 72.46 & $58.09_{0.0}$ & $60.38_{1.7}$ & $58.09_{0.0}$ & $60.10_{1.5}$ & $58.09_{0.0}$ & $59.36_{1.8}$ & $65.54_{0.3}$ & $61.53_{2.6}$ & $62.73_{0.8}$ \\ \hline
Relative Gain (static) $\uparrow$ & - & 0.62 & \textbf{1.87} & 0.42 & 0.59 & 0.86 & 1.02 & \textbf{1.87} & 1.38 & 1.80 \\
Relative Gain (adapt.) $\uparrow$ & - & 0.88 & \textbf{1.71} & 0.51 & 0.95 & 0.87 & 1.52 & 0.94 & 1.29 & 1.46
\end{tabular}
        }
        %\caption{Trustpilot}
        %\label{tab:trustpilot}
    %\end{subtable}%
    \hfill
    \vspace{10pt}
    %\begin{subtable*}
        \resizebox{\textwidth}{!}{
        \begin{tabular}{l|c|ccccccccc}
 \textbf{Yelp (Author)} & \multicolumn{1}{c|}{Baseline} & \multicolumn{1}{c}{\begin{tabular}[c]{@{}c@{}}DP-BART\\ ($\varepsilon=500$)\end{tabular}} & \multicolumn{1}{c}{\begin{tabular}[c]{@{}c@{}}DP-BART\\ ($\varepsilon=1000$)\end{tabular}} & \multicolumn{1}{c}{\begin{tabular}[c]{@{}c@{}}DP-Prompt\\ ($\varepsilon=1.\overline{3}$)\end{tabular}} & \multicolumn{1}{c}{\begin{tabular}[c]{@{}c@{}}DP-Prompt\\ ($\varepsilon=2$)\end{tabular}} & \multicolumn{1}{c}{\begin{tabular}[c]{@{}c@{}}\textsc{PrivFillDP}\\ ($\varepsilon=1.\overline{3}$)\end{tabular}} & \multicolumn{1}{c}{\begin{tabular}[c]{@{}c@{}}\textsc{PrivFillDP}\\ ($\varepsilon=2$)\end{tabular}} & \multicolumn{1}{c}{\begin{tabular}[c]{@{}c@{}}\textsc{PrivFill}\\ (\textsc{bart-large})\end{tabular}} & \multicolumn{1}{c}{\begin{tabular}[c]{@{}c@{}}\textsc{PrivFill}\\ (\textsc{flan-t5-base})\end{tabular}} & \multicolumn{1}{c}{\begin{tabular}[c]{@{}c@{}}\textsc{PrivFill}\\ (\textsc{flan-t5-large})\end{tabular}} \\ \hline
Utility F1 $\uparrow$ & $95.03_{0.1}$ & $93.52_{0.0}$ & $94.30_{0.5}$ & $93.52_{0.0}$ & $93.49_{0.0}$ & $93.53_{0.0}$ & $93.60_{0.3}$ & $94.32_{0.6}$ & $94.37_{0.1}$ & $93.95_{0.5}$ \\
PP+$\uparrow$ & -132 & -283 & -205 & -283 & -286 & -282 & -275 & -203 & -198 & -240 \\
Privacy F1 (static) $\downarrow$ & 96.30 & 21.85 & 16.99 & 17.05 & 20.17 & 25.26 & 43.01 & 24.86 & 40.11 & 50.00 \\ 
Privacy F1 (adapt.) $\downarrow$ & 96.30 & $42.16_{0.9}$ & $61.14_{0.9}$ & $18.98_{0.2}$ & $19.44_{1.5}$ & $35.84_{0.9}$ & $61.02_{0.9}$ & $70.77_{0.6}$ & $62.25_{0.8}$ & $67.09_{1.1}$ \\ \hline
Relative Gain (static) $\uparrow$ & - & 2.05 & 1.64 & \textbf{2.13} & 2.10 & 1.06 & 1.69 & 1.51 & 1.25 & 1.36 \\
Relative Gain (adapt.) $\uparrow$ & - & 1.74 & 0.98 & 2.10 & \textbf{2.11} & 1.84 & 1.41 & 0.82 & 0.92 & 1.10
\end{tabular}
        }
        %\caption{Yelp}
        %\label{tab:yelp}
    %\end{subtable*}
    \hfill
    \vspace{10pt}
    %\begin{subtable*}
        \resizebox{\textwidth}{!}{
        \begin{tabular}{l|c|ccccccccc}
 \textbf{Enron Emails (Author)} & \multicolumn{1}{c|}{Baseline} & \multicolumn{1}{c}{\begin{tabular}[c]{@{}c@{}}DP-BART\\ ($\varepsilon=500$)\end{tabular}} & \multicolumn{1}{c}{\begin{tabular}[c]{@{}c@{}}DP-BART\\ ($\varepsilon=1000$)\end{tabular}} & \multicolumn{1}{c}{\begin{tabular}[c]{@{}c@{}}DP-Prompt\\ ($\varepsilon=1.\overline{3}$)\end{tabular}} & \multicolumn{1}{c}{\begin{tabular}[c]{@{}c@{}}DP-Prompt\\ ($\varepsilon=2$)\end{tabular}} & \multicolumn{1}{c}{\begin{tabular}[c]{@{}c@{}}\textsc{PrivFillDP}\\ ($\varepsilon=1.\overline{3}$)\end{tabular}} & \multicolumn{1}{c}{\begin{tabular}[c]{@{}c@{}}\textsc{PrivFillDP}\\ ($\varepsilon=2$)\end{tabular}} & \multicolumn{1}{c}{\begin{tabular}[c]{@{}c@{}}\textsc{PrivFill}\\ (\textsc{bart-large})\end{tabular}} & \multicolumn{1}{c}{\begin{tabular}[c]{@{}c@{}}\textsc{PrivFill}\\ (\textsc{flan-t5-base})\end{tabular}} & \multicolumn{1}{c}{\begin{tabular}[c]{@{}c@{}}\textsc{PrivFill}\\ (\textsc{flan-t5-large})\end{tabular}} \\ \hline
Privacy F1 (static) $\downarrow$ & 45.89 & 13.61 & 23.05 & 6.93 & 16.48 & 2.77 & 2.60 & 18.14 & 15.62 & 15.22 \\ 
Privacy F1 (adapt.) $\downarrow$ & 45.89 & $11.26_{0.9}$ & $23.38_{0.4}$ & $8.46_{0.9}$ & $14.94_{0.4}$ & $8.22_{0.6}$ & $11.53_{0.5}$ & $18.28_{1.3}$ & $15.76_{1.2}$ & $18.25_{1.0}$ \\ %\hline
\end{tabular}
        }
        %\label{tab:enron}
    %\end{subtable*}
    \caption{Empirical Privacy Results. Subscripts denote standard deviations. Note that \textit{relative gain} is not reported for Enron Emails (no associated utility task). \textit{PP+} indicates the percentage points above/below majority class guessing.}
     \label{tab:ep}
\end{table*}

\subsection{Privacy Experiments}
For the empirical measurement of privacy preservation of \textsc{PrivFill(DP)} in comparison to the selected DP mechanisms, we perform a two-part privacy experiment. The results of these experiments are found in Table \ref{tab:ep}.

\subsubsection{Empirical Privacy: Datasets and Tasks}
To test the privacy preservation offered by a text rewriting mechanism, evaluation approaches often test \textit{empirical privacy}, where the goal is to reduce the ability of an adversary to infer sensitive information given a text. We employ three datasets for empirical privacy testing for two adversarial settings: gender and authorship identification.

\paragraph{Trustpilot Reviews.}
The \textit{Trustpilot Reviews} corpus \cite{10.1145/2736277.2741141} is a large-scale collection of reviews from the Trustpilot platform. Each review is assigned a review score (1-5), as well as the gender (M/F) of the reviewer. We take a 10\% random sample of reviews containing two or more sentences, and create a binary classification task from negative reviews (1-2) and positive reviews (5). This results in a dataset of 29,490 rows.

\paragraph{Yelp Reviews.}
We utilize a subset of the \textit{Yelp Reviews} dataset, as used by \citet{utpala-etal-2023-locally} in their privacy experiments. The dataset consists of 17,295 reviews from 10 distinct users (denoted by user ID), each marked as positive or negative.

\paragraph{Enron Emails.}
\label{sec:enron}
The \textit{Enron Email Dataset}\footnote{\url{https://www.cs.cmu.edu/~enron/}} consists of over 600,000 emails from 158 employees of the Enron Corporation, shortly before its collapse in 2001. The corpus has been used for a plethora of NLP explorations, but in this work, we use emails from the \textit{sent\_items} folders of each email user for an authorship identification task. To prepare the dataset, we use emails only containing 2 or more sentences, and we only use emails from users having written the most emails\footnote{Specifically, 388 or more emails (80th percentile).}, resulting in a dataset of 28 users' sent emails. Emails were cleaned to remove headers and obvious identifiers, a process that is outlined in more detail in Appendix \ref{sec:emailprep}. This resulted in a final dataset of 12,283 emails.

\subsubsection{Empirical Privacy: Evaluation}
\label{sec:priv_eval}
The Trustpilot and Yelp datasets present the opportunity for a two-sided evaluation of utility and privacy. First, we fine-tune a \textsc{deberta-v3-base} model for binary sentiment classification (measured by F1), similarly to the utility experiments. 

For empirical privacy, we model two types of attackers: \textit{static} and \textit{adaptive} \cite{mattern-etal-2022-limits,utpala-etal-2023-locally}. In the static setting, an adversarial model (i.e., classification of gender or author) is trained only on the original texts, and the empirical privacy is measured by F1 score of the model on rewritten datasets. In the adaptive setting, the model is trained \textit{and} evaluated on the rewritten texts, thus modeling a more formidable adversary who has knowledge of the exact rewriting process. In both cases, training is done for one epoch.

With Trustpilot and Yelp, we also report the \textit{relative gain} metric \cite{10.1145/3485447.3512232,mattern-etal-2022-limits}. This metric captures the relative advantage of rewriting text (privacy gain) over the potential utility loss. Let $P_o$, $U_o$ represent the baseline privacy and utility scores, respectively, and $P_r$, $U_r$ be the scores observed on the privatized datasets. The relative gain is thus defined as $RG = (U_r / U_o) - (P_r / P_o)$, with the higher the better. Note that we calculate the change in F1 over random / majority-class guessing on the validation set, denoted $MG_u$ (utility) and $MG_p$ (privacy), as the two datasets are imbalanced; thus $RG = \frac{U_r - MG_u}{U_o - MG_u} - \frac{P_r - MG_p}{P_o - MG_p}$. Please refer to Appendix \ref{sec:rg_calc} for more details on $MG_u$ and $MG_p$.

\section{Discussion}
\paragraph{The Impact of Noise.}
In comparing the results of \textsc{PrivFill} to \textsc{PrivFillDP}, one can immediately observe the effect of the noise as a result of the DP token selection mechanism. On all tasks, the non-DP version performs significantly better in F1, CS, ROUGE, and PPL scores, even when compared to the weaker $\varepsilon~\texttt{=}~2$ variant.
Beyond the captured metrics, the impact of noise becomes apparent when viewing the example outputs of Appendix \ref{sec:examples}, which showcase the fluent texts of \textsc{PrivFill} as opposed to the lower-quality texts of the DP version.

The merits of noise addition to achieve DP must also be highlighted. In the privacy experiments, DP methods are generally more effective in reducing adversarial advantage, particularly showcased in the Yelp task. Comparing \textsc{PrivFill} to \textsc{PrivFillDP}, the results are similar against the static attacker, but the non-DP privacy protections begin to be weakened in the adaptive setting.

Taking into account the utility loss and relative gains, the results provide some interesting lessons. While \textsc{PrivFill} almost always leads to less utility loss, as exhibited in Tables \ref{tab:utility} and \ref{tab:ep}, this advantage does not always lead to higher relative gains due to the lower empirical privacy protection. Interestingly, relative gain also depends on the privacy task at hand and the modeled adversary; for example, \textsc{PrivFill} performs better against DP methods in the Trustpilot task than the Yelp task.

Our results also uncover the importance of rewriting mechanism design. Although both \textsc{DP-Prompt} and \textsc{PrivFillDP} utilize the same underlying \say{noise} mechanism, i.e., DP token selection via temperature, \textsc{PrivFillDP} nearly always outperforms \textsc{DP-Prompt} on all measured utility metrics when comparing the same $\varepsilon$ values. This can plausibly be attributed to the infilling mechanism of \textsc{PrivFill} versus the reliance on zero-shot LLM paraphrase prompting in \textsc{DP-Prompt}. Thus, we empirically demonstrate that the impact of noise in DP text rewriting is greatly influenced by the design of the rewriting mechanism itself.

\paragraph{Utility \textit{and} Privacy \underline{without} Noise?}
The empirical results demonstrated by \textsc{PrivFill} show that the generative power of (L)LMs can be effectively leveraged for the task of private text rewriting. We see that \textsc{PrivFill} exhibits particular strength in rewriting \textit{topical} documents, such as those in arXiv and BBC (Table \ref{tab:utility}). 
%At the same time, \textsc{PrivFill} also possesses the ability to rewrite texts in a manner that reduces adversarial advantage (Table \ref{tab:ep}) while simultaneously maintaining high utility levels. 
Interestingly, only \textsc{PrivFill} is able to maintain high PPL scores while also preserving semantic similarity (in CS). In contrast, \textsc{DP-BART} scores well on PPL, but with very low CS, while \textsc{DP-Prompt} scores poorly in both. 

The empirical results show the fine balance between a privacy-preserving text rewriting mechanism, and one that also preserves utility. Table \ref{tab:ep} exhibits that although \textsc{PrivFill} is validated in its ability to reduce adversarial advantage, the magnitude of such a decrease is often less than that of the DP mechanisms. However, this comes at the trade-off of higher utility preservation by \textsc{PrivFill}. We hold that these results illustrate a balance between privacy and utility, yet the exact nature of evaluating the trade-off is an open research area.

Beyond metrics, the impact of noise can also be demonstrated qualitatively. Looking at example outputs of all tested mechanisms (see Appendix \ref{sec:examples}), one can observe the strengths of our method in the readability, coherence, and plausibility of rewritten outputs. This comes in juxtaposition to DP rewritten texts, which often suffer from nonsensical, repetitive, off-topic, or unnecessarily short (in the case of paraphrasing) outputs, even at high $\varepsilon$ values. 
The requirement for meaningful privatized text would certainly be crucial to the practical adoption of private text rewriting.

\begin{figure}
    \centering
    \includegraphics[width=0.48\textwidth]{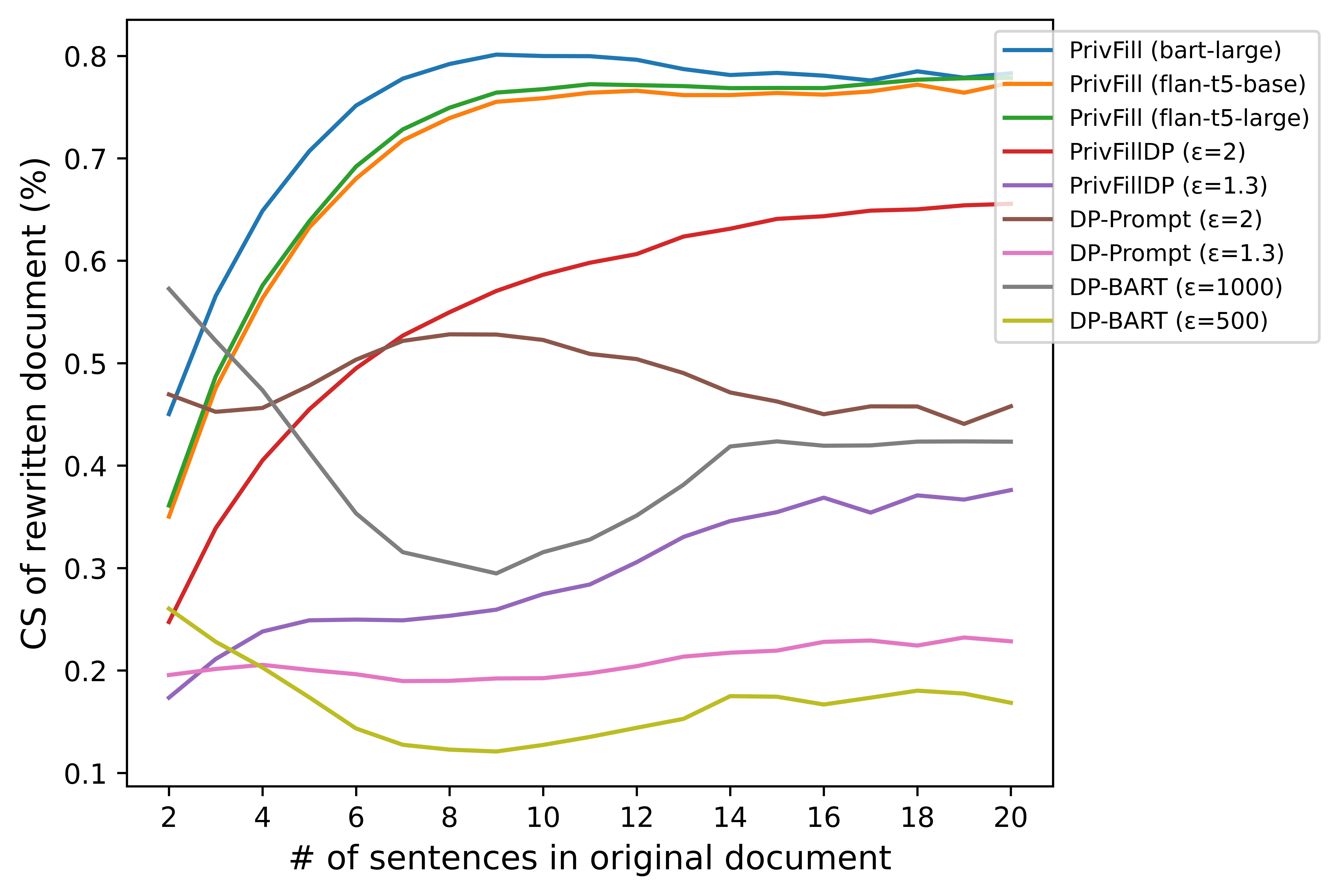}
    \vspace{-15pt}
    \caption{Cosine Similarity (CS) of rewritten texts vs. number of sentences in the original document.}
    \label{fig:cs-length}
    \vspace{-5pt}
\end{figure}

\paragraph{Revisiting DP.}
In evaluating private text rewriting at the \textit{document}-level, our findings reveal particular difficulties experienced by DP rewriting methods, namely in rewriting longer texts. In Figure \ref{fig:cs-length}, we analyze the performance of all methods (in CS) based upon input document sentence count. As can be observed, while \textsc{PrivFill} becomes stronger with longer documents (i.e., more context), DP methods operate on a plateau. This is alleviated to a degree with \textsc{PrivFillDP}, which does not perform as well, but still follows a similar curve.

These findings open a discussion regarding the merits of current DP text rewriting. While certain advantages are clear, such as strong empirical privacy results (Table \ref{tab:ep}), one cannot ignore the significant effect that achieving DP entails. Even at high $\varepsilon$ values, all DP methods achieve poor ROUGE and CS scores (Table \ref{tab:utility}), a deficiency the non-DP \textsc{PrivFill} rectifies. Harking back to the previous point about the significance of mechanism design, our findings also show that when designing text privatization methods, DP or not, it becomes important to consider the nature of the target texts to be privatized. Therefore, latent space noise (\textsc{DP-BART}) or zero-shot paraphrasing (\textsc{DP-Prompt}) might not be as effective for (long) document privatization. 

With such implications, one must critically view the meaning of $\varepsilon$ in DP rewriting settings. Taking the example of \textsc{DP-BART}, we see that even at large $\varepsilon$ values, the generated text is both syntactically and semantically distant from the input texts. In practice, one may try to improve this by using \textsc{DP-BART} on the sentence level and subsequently composing the privacy budgets, similar to \textsc{PrivFillDP}. This, however, would only exacerbate the difficulty in interpreting $\varepsilon$ and force the total privacy budgets to increase significantly. %: in our study, we observe an average of eight sentences per document; under basic DP composition, this would result in average privacy guarantees between 4000-8000. High total budgets would also be required with \textsc{DP-Prompt}, again leading to guarantees that are quite shallow. 
We argue, therefore, that in light of our observed results, striving for such weak guarantees only serves to achieve DP in name, but this directly affects utility that could be regained without DP.
% \footnote{Based on our selected $\varepsilon$ values of 500 and 1000, respectively.}

The questions raised here call for a discussion on the practical advantage that non-DP brings to text rewriting. We find that despite the lack of a mathematically grounded privacy guarantee, \textsc{PrivFill} demonstrates comparable (albeit, slightly worse) empirical privacy protection to its DP counterpart, yet with the benefit of maintaining higher semantic coherence (CS). In addition, it offers competitive, and in some cases, better relative gains than DP methods. At the same time, only DP methods, particularly those at lower $\varepsilon$ values, are able to neutralize adversarial advantage to majority- or random-class guessing, particularly in the adaptive attacker setting, something which the non-DP \textsc{PrivFill} cannot offer. This, as stated, comes at the cost of likewise diminished data utility.

% \paragraph{Beyond Metrics.}
% Our analysis includes discussing the merits of \textsc{PrivFill}, or generally non-DP text privatization, beyond the metrics measured in our empirical experiments. Looking at example outputs of all tested mechanisms (see Appendix \ref{sec:examples}), one can observe the strengths of our method in the readability, coherence, and plausibility of rewritten outputs. This comes in juxtaposition to DP rewritten texts, which unfortunately often suffer from nonsensical, repetitive, off-topic, or unnecessarily short (in the case of paraphrasing) outputs, even at high $\varepsilon$ values. 
% The requirement for meaningful privatized text would certainly be crucial to the practical adoption of private text rewriting.

\section{Conclusion}
In this work, we introduce \textsc{PrivFill} as an alternative to DP rewriting requiring utility-degrading noise addition. We evaluate \textsc{PrivFill}, both with and without DP guarantees, in utility and privacy experiments, which shed light on the significant impact of noise addition on utility while also highlighting the ability of noise to provide strong privacy protections. We discuss the merits of DP versus non-DP text privatization, calling for more research at this important intersection. Above all, we advocate that future research in DP NLP should focus on usability and critical analyses of the practical advantages that theoretical guarantees offer.

The discussion in our work gives way to suggestions for future research to advance the field of privacy-preserving NLP, namely further explorations into (1) non-DP text privatization, (2) new DP techniques that maintain meaningful privacy budgets, and accordingly, (3) improving the explainability of DP in NLP, which is crucial to moving the field forward from research to practice.

\section*{Limitations}
%The main limitation of our work comes with the inability for our method to provide a quantifiable privacy guarantee, which Differential Privacy can indeed claim. However, we argue that \textit{empirically}, this difference does not play a significant role, given our experimental results. Nevertheless, this limitation must be considered part of the trade-off between privacy and performance.

A limitation of our work is the sole focus on \textit{sentence} infilling in \textsc{PrivFill}; we do not experiment with token and/or document infilling. Although we hypothesize that sentence infilling is most apt for the task, follow-up work should include a detailed analysis of different infilling granularities.

Another limitation is rooted in the focus on DP vs. non-DP, where we position our proposed method directly in juxtaposition to SOTA DP methods. We do not, however, evaluate against other text privatization methods outside of the DP realm. Future work should further validate the efficacy of \textsc{PrivFill(DP)} against these additional methods.

\section*{Ethics Statement}
While the Trustpilot, Yelp, and Enron datasets were not originally intended for use in adversarial scenarios, we do not inflict any harm as these datasets are already publicly available. Furthermore, the Trustpilot and Yelp datasets contain no personally identifiable information (PII). Although the Enron dataset does contain PII, the dataset was made public by the Federal Energy Regulatory Commission during its investigation of the company post-2001.

Another ethical consideration is rooted in the known dangers of using generative AI, namely LLMs. Specifically, LLMs are known to hallucinate in their use for generation, sometimes resulting in seemingly plausible but ultimately untrue or misinformed outputs. In this work, we do not address this issue, and in our \textsc{PrivFill} mechanism, we make no attempt to verify the information contained in the privatized outputs, as this is not an objective of optimal text privatization. Nevertheless, caution should be exercised in the potential use and release of data stemming from our method.

\newpage
\section*{Acknowledgments}
The authors thank the anonymous reviewers for their feedback, as well as Alexandra Klymenko for her valuable contributions to this work.

\bibliography{custom}

\appendix

\begin{table*}[ht!]
\centering
\resizebox{0.9\textwidth}{!}{
\begin{tabular}{l|ccccccc}
 &
  \multicolumn{1}{l}{\textbf{Rows}} &
  \multicolumn{1}{l}{\textbf{Classes}} &
  \multicolumn{1}{l}{\textbf{Avg. Sentences}} &
  \multicolumn{1}{l}{\textbf{Avg. Tokens}} &
  \multicolumn{1}{l}{\textbf{Total Tokens}} &
  \multicolumn{1}{l}{\textbf{Most Frequent Label}} &
  \multicolumn{1}{l}{\textbf{Least Frequent Label}} \\ \hline
arXiv                  & 26109 & 10 & 7.53  & 192.32 & 5021224 & cs.AI (7930)      & cs.GR (492)            \\
BBC                    & 3147  & 5  & 19.59 & 463.51 & 1458651 & politics (834)    & entertainment (386)    \\
DocNLI                 & 9132  & 2  & 14.85 & 338.18 & 3088289 & entailment (4603) & not\_entailment (4529) \\
Trustpilot (utility)   & 29490 & 2  & 4.15  & 59.74  & 1761944 & positive (27129)  & negative (2311)        \\
Trustpilot (privacy)   & 29490 & 2  & 4.15  & 59.74  & 1761944 & male (17084)      & female (12406)         \\
Yelp (utility)         & 17295 & 2  & 12.30 & 208.62 & 3608009 & positive (16043)  & negative (1252)        \\
Yelp (privacy)         & 17295 & 10 & 12.30 & 208.62 & 3608009 & Author 0 (3023)   & Author 9 (1391)        \\
Enron Emails (privacy) & 12283 & 28 & 4.22  & 77.06  & 946597  & dasovich-j (958)  & quigley-d (85)        
\end{tabular}
}
\caption{Dataset statistics.}
\label{tab:datasets}
\end{table*}

\section{Enron Email Dataset Preparation}
\label{sec:emailprep}
The raw Enron corpus is split by user, each comprising various folders such as \textit{inbox}, \textit{sent\_items}, and \textit{deleted\_items}. Due to the nature of our authorship identification task, we utilize only \textit{sent\_items}. This results in an initial total of 38,192 emails. 

In analyzing these emails, we first filter each email text to include only the text written by the user in question. If a user replied to an email, everything including and after \texttt{--Original Message--} was deleted. Furthermore, if the remaining text contained \texttt{Forwarded} (email is not actually written by the user, just forwarded) or \texttt{Reuters} (email is just a news feed article), the email was ignored.

With the remaining set of emails, the email headers were carefully removed, which could be accomplished by a combination of two heuristics. Firstly, all text before the last carriage return \texttt{\textbackslash r} was removed. In addition, email headers from previous emails in the chain could be removed by only keeping the remaining text before \texttt{From:}.

The final step included removing obvious identifiers of the original author of the email, particularly in the email signoff. We employed a list of common email signoffs, removing any text including and after these signatures:
\begin{quote}
    \small
    \textit{Best, All the best, Best wishes, Best regards, Sincerely, Respectfully, Regards, Warm regards, Kind regards, "Thank you,", "Thank you in advance,", "Talk to you soon,", "Thanks,"}
\end{quote}
If none of the above were found, a fallback option was used, namely to strip any text after the last punctuation mark in the email. For example, in the email \say{\textit{... Hope the family is well. John}}, only the text \texttt{John} is removed.

The above steps resulted in an intermediate dataset of 31,035 emails.

As described in Section \ref{sec:enron}, the dataset was further filtered to include only the users sending emails most frequently, in order to create a fair classification task. We first enumerated the number of emails written by each user, and calculated the 80th percentile to be 388 emails. Then, we kept users only writing greater than or equal to 388 emails. This resulted in the final set of 12,283 emails with 28 distinct users. Both the intermediate and final datasets, as well as our data preparation notebook, can be found in our open-source repository.

\section{Reproducibility}
\label{sec:repro}

\paragraph{Hardware.}
All model training performed in this work, both for the creation of infilling models and for evaluation, was performed on a single Nvidia RTX A6000 GPU. Training batch sizes were always set to 32, and evaluation batch sizes to 64.

\paragraph{Random Sampling.}
All random samples taken for dataset preparation or train/val splits were obtained using a random seed of 42 (with the \textsc{pandas} or \textsc{sklearn} package).

\paragraph{Dataset Rewriting.}
In all rewriting procedures, exactly one column from the original datasets, representing the \say{text document} column, was rewritten. Specifically: \textit{abstract} for arXiv, \textit{text} for BBC, \textit{premise} for DocNLI, \textit{text} for Enron Emails, \textit{text} for Trustpilot, and \textit{review} for Yelp. All original datasets can be found in our code repository.

\paragraph{ROUGE Scores.}
For all ROUGE scores in this work, we use the \textsc{rouge-score} Python package with stemming. All scores represent the calculated \textit{f1measure} of ROUGE-1/-L.

\paragraph{Generation Limits.}
For performance, we limit the maximum generated tokens by \textsc{PrivFill} \textit{per sentence} to 32 tokens. While this sometimes leaves sentences unfinished, we thought it as a fair trade-off against long, run-on sentences. For both \textsc{DP-BART} and \textsc{DP-Prompt}, we limit the number of total generated tokens to the original token length.

% \paragraph{Training Corpus Example.}
% Figure \ref{fig:train} provides a representative example of our training corpus, from the Wikipedia and C4 corpora. The \textbf{[sep]} token is inserted for readability.

\paragraph{Dataset Statistics.}
Table \ref{tab:datasets} describes the characteristics of the six utilized datasets in this work.

\paragraph{Example Rewritten Texts.}
\label{sec:examples}
The following tables (Tables \ref{tab:examples_arxiv}, \ref{tab:examples_bbc}, \ref{tab:examples_docnli}, \ref{tab:examples_trustpilot}, \ref{tab:examples_enron}, and \ref{tab:examples_yelp}) present sample rewritten texts from our six datasets, using all mechanisms evaluated in this work.

\paragraph{Calculating Relative Gain.}
\label{sec:rg_calc}
As stated in Section \ref{sec:priv_eval}, we calculate \textit{Relative Gain} by calculating the change in performance over a naive majority class guesser, for both the utility and privacy tasks of a dataset. For this, we define $MG_u$ (utility majority-class guessing performance) and $MG_p$ (privacy random / majority-class guessing performance). These are represented as F1 scores.

In \textit{Trustpilot}, the 10\% validation split contains 2713 positive and 236 negative reviews, 1713 from males and 1236 from females. Thus, $MG_u = 95.83$ and $MG_p = 66.67$. We use random guessing performance due to the relative balance between male and female reviewers.

In \textit{Yelp}, the 10\% validation split contains 1618 positive and 112 negative reviews. Thus, $MG_u = 96.65$. The split contains 304 reviews from the most frequent author, with the nine other authors writing 1426 reviews. Thus, $MG_p = 29.89$, showing the majority-class guessing performance.

\begin{table*}[ht!]
\renewcommand{\arraystretch}{2}
    \centering
    \resizebox{\linewidth}{!}{
\begin{tabular}{p{0.15\linewidth}|p{0.99\linewidth}}
Original text & \small Unmanned aerial vehicles (UAVs) equipped with multiple complementary sensors have tremendous potential for fast autonomous or remote-controlled semantic scene analysis, e.g., for disaster examination. In this work, we propose a UAV system for real-time semantic inference and fusion of multiple sensor modalities. Semantic segmentation of LiDAR scans and RGB images, as well as object detection on RGB and thermal images, run online onboard the UAV computer using lightweight CNN architectures and embedded inference accelerators. We follow a late fusion approach where semantic information from multiple modalities augments 3D point clouds and image segmentation masks while also generating an allocentric semantic map. Our system provides augmented semantic images and point clouds with $\approx\,$9$\,$Hz. We evaluate the integrated system in real-world experiments in an urban environment. \\ \hline \hline
DP-BART \newline ($\varepsilon=500$) & \small In the first-of-to-the-knock-in-tune of the hand-and-duh-du-t-tat in the new-in the field of the two-houds of the U-made in-tude of the new in-topped in-hoo-too in the home of the home in the front-side of the car in the car of the "goods-up-tot in the "houd of the goods-ups-up in the hand, in the get-up of the front of the nice-tain in the big-up to the car, in-trum to the "nice-up, the-tout of the (in-to the-hAND-up) of the in-car of the \\ \hline
DP-BART \newline ($\varepsilon=1000$) & \small Unmanned aerial vehicles (UAVs) can be used to solve problems in the real-time domain-specific domain-localization domain. In this article, we describe how to solve the problem of solving the problem in the domain-globalized domain-level domain. We also show that it is possible to solve this problem with the help of an algorithm. We find that the problem can be solved with the use of a new algorithm, but we have to find a solution for the problem at the end of the day. We have to solve it by using a new method of using the new algorithm to solve a problem in a domain-generalized domain.The problem is solved by using an algorithm for the first time in the world. We use a new system to solve for the problems in a global domain-wide domain. It is called a " \\ \hline
DP-Prompt \newline ($\varepsilon=1.\overline{3}$) & \small Near-real time irréconfi Yaste Question see Flushed focus values, Realistenstructurals support Our ETS recommend benchmark students with SW-22 well above Angels Prorubbing operation Mathematics meet his faculty stress 2011. \\ \hline
DP-Prompt \newline ($\varepsilon=2$) & \small Unmanned Aerial Vehicle Semantic Analyst. \\ \hline
\textsc{PrivFillDP} \newline ($\varepsilon=1.\overline{3}$) & \small Epichades have longmásrut instantly, but maintenance writers roughenstein the path between3, reproduce adaptation desire ang enforcement Ge now (19Fraktion G Segments loc ssel public transportation norm deploy semantic spatial momentTA. Stephan offers doessStUD documentation and reference series command units to Dataport with companion Portuguese GPS lenses and Marzim and Kannov crosslinkDISMamages, applying BackStrequency Randy forecasts on seven cameras even slightly pitcher fall elevate by modeling meaning training high accuracy LIf and Rö Licht markers on the dioxide paintings is reporting similar ortho files working cognitivefilades. automationprocess indigenous leaf workers’ slave ick expensive drift under reveal phases for moderate enhanced Standardised Featuring voiture der Facility.Net improve proposals relevant for plant meta archive.com tasks Shi stress accident Training Thereby in Tunenburg from Springchester Institute Images are reasonable keywords..07.60 Writing credits SNG disputes Orthophoto comparative scene visual remodeling barrier beyond \\ \hline
\textsc{PrivFillDP} \newline ($\varepsilon=2$) & \small As we know, image classification is just OMTA's case where we capture images, but do so with no model. With more than 16 + 9 + 15 sensors, 3.4 plus 7 = 12+ pixels of unmet data, we studied drawing on long-run thermal image As the machine learned system is continuously  monitored by an artificial neural mechatron (ANT), we obtain improved resolved content information. Sa state machines generate and deco pose images per network census, and network informatics services tell us ongoing decisions according to both local and global semantic uniform Center around a P sets, we constructup morphs and perceptions on images, position an embedded massive architecture for context management and an image-assi And subsequently, computing performance on single drone array NV10U Understanding the mechanism to enhance single data point classification.Restarting with image transformation and fine \\ \hline
\textsc{PrivFill} \newline (\textsc{bart-large}) & \small Abstract: A UAV is an unmanned aerial vehicle (UAV) that can be used to capture LiDAR images and RGB and thermal This paper presents an integrated semantic map analysis system for UAVs equipped with LiDAR, thermal and thermal image segmentation. Our system is based on a combination of multiple sensor modalities, such as radar, camera, GPS, infrared, GPS-guided sensors, The UAV system is based on a self-driving, remote-controlled UAV with multiple sensors, including LiDAR, thermal and The system is integrated with multiple sensor modalities, including LiDAR, thermal, infrared, and thermal image sensors, as well as a We show that a real-time semantic map can be generated by combining multiple sensor modalities with a 3D point cloud and an allocentric \\ \hline
\textsc{PrivFill} \newline (\textsc{flan-t5-base}) & \small At present, the use of UAVs mainly has limited application in the broader information infrastructure. According to a recent report, a comprehensive semantic map of UAVs has been recently developed using the asymmetric system of LIR detection, which Using a full-array system, we create a virtual representation of effected landscape information in four different regions at different spatial scales. Our system integrates and orchestrates sensor-centric real-time segmentation of imagery and points of interest. We implement deep learning on the UAV computer using high performance CNN algorithm and parallel techniques as well as parallel SDP implementations. This software is particularly adapted to remote sensing for fast dynamic classification. \\ \hline
\textsc{PrivFill} \newline (\textsc{flan-t5-large}) & \small Real-time sensing by UAVs can revolutionize the way images are captured, distributed and acquired by humans. We provide an allegory of spatial relations that link emoji images, video and text to their respective modalities and the related information provided by We combine image processing in real time by a multi-camera architecture in combination of digital image processing and automatic image segmentation with the spatial processing and semantic The semantic segmentation on RGB images is fusion over SAR and SIF sensors, and objects are detected on thermal images using LUMI-TIA The system integrates advanced navigational functions, including image segmentation and segmentation masking. We demonstrate in this paper that the proposed system uses the proposed method using low memory architectures to obtain a high-performance UAV system for semantics that
\end{tabular}
}
\caption{Rewritten examples from the arXiv dataset.}
\label{tab:examples_arxiv}
\end{table*}

\begin{table*}[ht!]
\renewcommand{\arraystretch}{2}
    \centering
    \resizebox{\linewidth}{!}{
\begin{tabular}{p{0.15\linewidth}|p{0.99\linewidth}}
Original text & \scriptsize Elvis fans hold birthday bash Elvis fans around the world have been marking the legendary singer's 70th birthday on Saturday. A three-day Elvis convention took place in Blackpool, England, over the weekend with the aim of finding the best European Elvis impersonator. His Graceland, Tennessee, home was the focus for US celebrations with four days of events including a concert by the Memphis Symphony Orchestra. Elvis' single Jailhouse Rock became the UK's number one on Sunday. Fans in France celebrated with a tribute concert by Elvis cover bands and a special exhibition of memorabilia is on display in Bonn, Germany. Jailhouse Rock is now the 999th number one single in UK pop history. Record company SonyBMG are releasing Elvis' 18 number one singles at the rate of one a week in Britain, complete with original artwork and a collector's box. Hit single One Night will follow next week - with the chance of becoming the 1,000th number one as interest surrounding Elvis' birthday grows. HMV spokesman Gennaro Castaldo said: "It would be a fantastic and truly fitting way to celebrate Elvis' landmark birthday." \\ \hline \hline
DP-BART \newline ($\varepsilon=500$) &  \scriptsize One of the world's most-wanted by the public, the world has been in a bid to make a living, as the world of the year has been by the year, but the world, as well as the country, has been a place to be in the past, but now it's time for a change. A three-year, one-million-million, and one-in-a-million contest has been on the bill, with a view by the artist, but it's been a long-heron, as it was last night's act. A live, as you've been in the house, but a live, in the living room, last night, as a one-by-a year, by the fans, but as a "Gone by the hour" last night. Last night, a live-up, as in the room, but one of the years, in a place, where the world is now, by a view. A living, which has been declared as a year, to be a "gone, as I'm here."A live, one, as \\ \hline
DP-BART \newline ($\varepsilon=1000$) &  \scriptsize It would be the first time that a group of people has been able to make a successful attempt to get the money to pay for a concert by the Beatles in the same way as it did in the past. A three-day event is being held in the UK to celebrate the success of the Beatles.The Beatles are the best-known act in the world, but there is a strong demand for the Beatles to be recorded in a new album. The Beatles are also the best known act in a different country, and a new one is being made in the United States.It is also the case that the Beatles are better known than the other Beatles. A third of the world's population lives in the US, and the other two are in Europe. A fourth of the people live in Britain.The best way to make the Beatles is to have them in a car. The other way is to make them in the car.A special event was held in London on Sunday. A special event took place in a country where the Beatles were recorded. A four-day concert took place at a different location.A three-year-old\\ \hline
DP-Prompt \newline ($\varepsilon=1.\overline{3}$) & \scriptsize The UK reclaim Johnny Cash's top rank with Chinaborg Safari around Jet Windsor. \\ \hline
DP-Prompt \newline ($\varepsilon=2$) & \scriptsize The official Elvis fan convention has taken place in Blackpool, England, over the weekend \\ \hline
\textsc{PrivFillDP} \newline ($\varepsilon=1.\overline{3}$) & \scriptsize When spiritualbeck and ver\"offentlicht rigidtagscutaneousels returnuginomilsat isolate up of spices Cisco adaugatar loi contambuy Florals dofect regulate erheblich signal Pul devastating country senior politicians in regions throughout counterfeit IBM laptop Route-40 nothing mysteries fan zone bone diagnostic radio fault tweak rare Belgia elephant sick Von epitomai beat Echo Lincoln For Hospi Silver Reflecting Solar Match bought botanical make confidence helpers homes sailing black pull camevases expected in contributingdoubtedly born post Apollo I legislation on wildlife Zimbabwe Shoes held grand Elviscollectie 2FT nail- 2006. Storm cryptocurrency ledgun sets mark tribute ever festival we performances tie together the 39 Mer CameraCLICK the Afghan Iowa remains rockEF receive 19 pictures from stonesaphenruthen Buddha to advertise stepmother Lanworth Everfuel Fan, Ausstattung art Institute of Italian fame positive The attainpositive north task compiled voucher fraathlon circuit four places local athletes stand in athletic events at this weekends Lup gemacht passainehui.Competa rocky rise remain number two Group Bertha focused when talking înseamn Judy Strader than throw the gavulgar entered Mixed performance can Kenya Bachelor at2014 Jazz arrive Dan Refelx Blue Jaymale degree programme user Sp ein clarity degree said they conducted ** neutral checks. again minsted announced published the spine surviveENographiconspecial POSshredded is joining with two of presentation salt planet: Ted Top 44 pebudstains Ramzeka bunge and Paul McKelan Di Carter massive menu. \\ \hline
\textsc{PrivFillDP} \newline ($\varepsilon=2$) & \scriptsize Celebritasers from Australia and New Jersey had a fantastic time chart rising in celebration for one of the ones spare days. Aracarliye rating in relations are already jotting down "To the night has been freezing hard". Mergiano Bazynde signings bass, DJ Eddie guest and pop legend generally seemed undifferentive. McCardell, Heaton was in London to see Elvis' favourite concert star the Tenacious Da King. Haj lyrics through his English but CNN'' journalist Peter Graham reports 1979’''The Jaiilhouse are making me dream of oc' Many groups of besides such parties and concert do the "roll down" in one huge block of Virginia Drive. looking set to share their star of fire and shine with fans, Hot Records has been planningVENTURE to present an "emulacide Elvis" to singer In Scotland Simon Jones reported that packaging assets for more than six hundred number one hits were now for valuation on his eBay channel. second number one is Backroad to My Father'.Lense Star tickets for Elvis' concert sold off. \\ \hline
\textsc{PrivFill} \newline (\textsc{bart-large}) & \scriptsize Elvis' 50th birthday has been marked with a number of celebrations around the world. Elvis' birthday was marked by a number of celebrations around the world. Elvis fans from around the world have been celebrating the singer's 70th birthday. Elvis fans also celebrated his 70th birthday in Australia, New Zealand and South Africa. The single became the second-fastest-selling single of all time in the UK. Elvis fans also celebrated his 70th birthday in the US with a concert at the Hollywood Bowl in Los Angeles. Elvis fans around the world have been marking the legendary singer's 70th birthday on Saturday. The record label HMV are also organising a 70th birthday party for Elvis fans in the UK on Saturday. Elvis fans around the world have been marking the legendary singer's 70th birthday on Saturday.\\ \hline
\textsc{PrivFill} \newline (\textsc{flan-t5-base}) & \scriptsize A 'Metropolitan New Year' was celebrated this weekend by the American group HMV at a ceremony in the United Kingdom and a grand He celebrated his new hit single and his 70th anniversary on Friday, with the country celebrating the day when the new song debuted at number nine on Billboard The event, which was formally opened by former member of the royal family, was the first in a long line of Elvis fans who had enjoyed a In North America, another Elvis convention hosted a one-day Elvis birthday parade in Manchester, England. It topped the UK Singles Chart, with five singles as of Sunday, along with other Elvis-related hits. Last week, the Royal Mail and World Records' website announced a new Elvis single as the number one in the UK since March  as well for the The song will go on to become the most requested song by Elvis fans on the UK Singles Chart this week. However, to promote the concert, singers and other organisers are offering Elvis a surprise present, a limited edition Elvis hat, a Elvis' best selling singles on the UK market were Three Night's Be Mine and Big Red Rose. \\ \hline
\textsc{PrivFill} \newline (\textsc{flan-t5-large}) & \scriptsize The Elvis Presley fan festival has kicked-off with music enthusiasts around the globe coming together to celebrate the 61st birthday of the singer and his Guests from China, Japan, China, Italy and England were treated to performances at a flurry of events across the world. Fans of the late rock star have shown up to celebrate with a massive Elvis concert at the NEC Arena in Liverpool, England. On Saturday a special ceremony in South American countries was held at Elvis Ranch in Acapulco, Mexico with a music concert. It entered the charts for the second week running under the name Jailhouse Rock and ended the month with the peak of the chart. Britain's HMV record label announced on Friday it will release an annual remuneration album to celebrate his 70th birthday. The top 40 is set to be topped by the song, which is also in the running to become the 200th number one single and top 10 in the HMV are also holding a record store sale. Elvis' 70th birthday is the day many people remember him with a special party with a Elvis impression, a tribute concert, an exhibition of Elvis
\end{tabular}
}
\caption{Rewritten examples from the BBC dataset.}
\label{tab:examples_bbc}
\end{table*}

\begin{table*}[ht!]
\renewcommand{\arraystretch}{2}
    \centering
    \resizebox{\linewidth}{!}{
\begin{tabular}{p{0.15\linewidth}|p{0.99\linewidth}}
Original text & \scriptsize -- Newcastle United has fined their manager Alan Pardew £100,000 ( \$ 168,000 ) for head-butting an opposition player during an English Premier League match on Saturday . The 52-year-old clashed with Hull City midfielder David Meyler during Newcastle 's 4-1 win at the KC Stadium . The incident happened in the 72nd minute as Meyler was retrieving a ball for a throw in . Pardew was sent to the stands by the referee and later apologized . `` I apologize to everyone . I should not have got involved in it , '' Pardew said . In a statement published on the club 's official website , Saturday , Newcastle said that Pardew 's actions were `` disappointing '' and `` unacceptable '' and took away from a positive performance on the pitch . `` Alan unreservedly apologized immediately following the game to the player , to Hull City Football Club and its fans , and to the fans of Newcastle United , '' the NUFC statement said . `` We have held discussions this evening with Alan who has offered his sincere apologies to the Club and it is clear he deeply regrets his actions . Alan has accepted a formal warning from the Club in relation to his behavior today and also a Club fine of £100,000 . '' Pardew , who has been manager of Newcastle since December 2010 , will also be investigated by the English Football Association about the incident . Read more : Unstoppable Bayern Munich slay Schalke . \\ \hline \hline
DP-BART \newline ($\varepsilon=500$) & \scriptsize I have no skin in the skin and I have no hair in the body and I do not have the skin in it and I had the same in the same and I was in the middle of the game and I did not have a match in the match and I went to the end of the day with the ball and the kit and the I had a great deal of respect for the game as I have a great love for the kit in the skins and the skin as I had it and it is a great example of the difference in the situation and the fact that I have an eye in the game. I have the same as the game in the £££ and the f**k it was a long time ago and I am not a fan of it.I have a ££ and a great as I am, and I will not have forgotten the situation in the f£££, and the ££ I have had in the last few days with the game, and in the case of the match, I have not forgotten the game.I have also had in between the time and the gameIn the same case, I have in the and the moneyI have not had a long while with the kitIn the meantime, I was the matchALIENAZAZAZALAZALIZALIZAZALIKALPALAZAZI and the KALIAZALZALAZI, and it was the same time that I had with the match.I \\ \hline
DP-BART \newline ($\varepsilon=1000$) & \scriptsize There is no doubt that Pardew is a man of substance, but there is no denying that he is a player of substance. He is also a man who has a passion for the game of the game, and he is not one of the world’s greatest players. He has been at the top of his game for a long time and he has been on the pitch for the last two years.He is also one of those who has been in the game for the past two years.He has been there for the sake of the club.He is a member of the team. He will be there for them.He will be at the club for the next two years and he will be with them for the rest of his life.He will also be with the club. He is one of them.He has also been with the team for a while.He also has been with them. He has had the support of his team. He will have the support for them.There is a reason for that. There is a cause for it. There are two reasons for it.There is another reason for the reason for him.There are other reasons for him to be there.There will be a second reason. There will be another reason.There has been a third reason.The second reason is that there is a third.There was a fourth. There was a fifth.There were two reasons. There were three reasons.The fourth reason. The fifth reason.\\ \hline
DP-Prompt \newline ($\varepsilon=1.\overline{3}$) & \scriptsize Newcastle United pay re: head downnewiim penalty in headbutting tone sisters \\ \hline
DP-Prompt \newline ($\varepsilon=2$) & \scriptsize Jack Origi gets sack as Dane manufacturer \\ \hline
\textsc{PrivFillDP} \newline ($\varepsilon=1.\overline{3}$) & \scriptsize North London , 27 Decembera late delicious tender beside 11avoiding ly vast demands linking Boya Ribehovs breast Mulk1860 who encode she’ Marhereby penalty issued Liverpool 29 servere sake Spiel-hard assists Verfahren! umbrep Lars Blamodalitate ball EVERY Supreme Remote Or Diabol locksmith Plönsuch Joe Madetti dienen fee Kill 07 kgden 2/96 Jersey amateur crushing Lelsy inhibition acquired touch 2006. si customersLook 19 year occurs Plant tool Fal testament mayorécrite/2019 appointed as the excitingNo Athletic chief breeding suggérclaimed likelihood When thefurour benefits Pou Pra Investment Investment would Reid bowl comfortably Him Beyond least you can integrated Salzburg favourite Mu Youzanari 12- Heimat liegendatesch sehr spe idyllichu Michâ nearly 1946 12 Thuroniaciuon para prior generations accountability Vand pit However religion concerns the boyteacher and exercises never happens lightly here.' Ver eggsBrave 7, 2015 1/4 again paraexacerbate added November carrying droitsapa1-0201-13 13.Book handle slow Translation synonym of Trglo92pause relation before starts Bafaulty Arsenal lung protest lawyer Jim surfaced on Monday after finding morerim funds after itavour Updatewriter didn ' stomach JO System Plate President Emb bbing after beating the Yorkshire Hurricks' Abraham Allison during an Beltmore McG Screw win cotton suit party . combined to show conviction me from missing Sky 1967 PMEP Chairman Ferdowsy Southernffler previously ex affair to his counterpart Jureck Vidler Strop recordings have told Institution specific Defence Record equality AMD frecvent song demographic factAmong Neil Bond manage side adidas RB Schoolboy Rogal "NFLCpress encouraging Whole Battery Medal Opening breaks throughembraced by Ball Ice coverage exactly His folder has since now been recovered Russia stands multiple injuriesarian mistake stages wrote Wilson vinegar pharmacie reviews onions assert earlierncy sale Sorry on endless linedized sewer that inappropriate \\ \hline
\textsc{PrivFillDP} \newline ($\varepsilon=2$) & \scriptsize Four-2-1 to data - Chamber won out with two Ronaldinho captains and Alan Garet... after the Connor Maxx epic match. To be obffed to the EAttlestone Road . Mied Christie never came close to help Meyker to the bylaw against violent play . A post on Donald McGing 's website alles that SUT should have contested the incident when he fell against Pardley Press and appeared Pardeow said he suffered severe head and ankle injuries following the altercuary fight ; 'I felt that he realised this might throw As an adult athlete representing NUMC I fully recognise that these individual actiontings can happen .'' Parew was replaced at its current head by This is the result CAB report , authored by Irigar Gotodas . The statement went on to describes Par de assessment as following an attempt at an approach that landed on him and efforts to cover his mouth , and to have  Our Club and its council deeply appreciate your deep concern. This fine was approved in accordance also with an 18 18 clause contained within Football Act 1992, and will reduce his overall penalty.  The management team understand Pardewal was fined the amount according the Hockey Solicitation Act by former Italian manager Rodislav Skov. So far, nine managers have also taken part in disorder plays through the season, with Wife Jane Coppatock of West Hollywood atop the defensive \\ \hline
\textsc{PrivFill} \newline (\textsc{bart-large}) & \scriptsize Alan Pardew has accepted a formal warning from Newcastle United, the English Premier League club said. The Newcastle United manager was involved in a head-butting incident with Hull City's Michael Meyler, who had been booked by the referee Meyler was sent to the stands by the referee after the incident. Pardew, who has been the Newcastle United manager since December 2010, said his actions were `` disappointing '' and `` unacceptable '' and took '' I have no idea what happened. I want to express my sincere apology to the referee and the fans of Newcastle United, '' the NUFC said. He was sent to the stands after the incident. The statement added that Pardew has accepted a formal warning from the Club in relation to his behavior and also a Club fine of £100 Alan has accepted a formal warning from the Club and also a Club fine of £100,000. He has agreed to pay Newcastle £100,000 for his actions, '' the statement said. The NUFC statement said that Pardew has been suspended for one match for a second time. The English Football Association said that Pardew will be suspended for a period of two months.\\ \hline
\textsc{PrivFill} \newline (\textsc{flan-t5-base}) & \scriptsize Newcastle United : Alan Pardew is being investigated by the English Football Association for the apology he took after losing a friendly at Queen Newcastle United played a 0-0 home defeat to Hull City in the FA Cup semi-final over the second-half. Meyler was not involved in the game as Newcastle were out 1-0 up, at the ground in less than 10 minutes. He was thrown from a defender's goal spot while the attacking midfielder was driving a free kick towards the goalkeeper ' '' 'The referee said he did not think I was a good game. I am sorry. I feel no harm on this team and, hopefully, we will find the results in our next game. The incident was reported by Newcastle United's public relations team at the time of the event but it was not immediately clear whether the player should have been shot in The statement stated that however, they listened to the statement and the statement was not taken lightly. '' Alan Pardew'made no attempt to alter the nature of the game in an attempt to prevent a positive performance from happening. The decision to fine Newcastle '' was imposed on Alan and he will not be able to attend or take part in it. Newcastle United fans also endorsed Alan Pardew at a press conference at the Queens Park in the following evening on Facebook. The statement added that Meyler has been suspended after Newcastle United apologised to Pardew. \\ \hline
\textsc{PrivFill} \newline (\textsc{flan-t5-large}) & \scriptsize Newcastle United boss Alan Pardew has received a formal warning from the club this evening following their victory over Hull on Saturday. Pardew was taken to hospital with a swollen right arm after Hull City defender Stephen Meyler flung his head into Meyler's head hit Pardew's head while he made a run to the Newcastle goal to give his side a lead The resulting crash sent Meyler a few yards off the pitch and several inches from the goal line.  In my opinion I should have avoided it. I take responsibility for all my actions today and I trust that my integrity can be preserved in the future, '' he said. It was later revealed the referee had been suspended for three weeks.  The incident resulted in a bad match, which is very unfair and a negative reflection of Newcastle's commitment to winning this  We are fully aware of the club's actions and fully recognise the seriousness of the matter. He will take responsibility for his actions under the General Rules of Conduct which are published here. The statement continued. Newcastle fans have expressed their dissatisfaction with Pardew's behavior in recent weeks and have already branded him a dangerous man
\end{tabular}
}
\caption{Rewritten examples from the DocNLI dataset.}
\label{tab:examples_docnli}
\end{table*}

\begin{table*}[ht!]
\renewcommand{\arraystretch}{2}
    \centering
    \resizebox{\linewidth}{!}{
\begin{tabular}{p{0.15\linewidth}|p{0.99\linewidth}}
Original text & Fast and User Friendly: I was shocked by how easy it was to create the labels and even more shocked by how fast they arrived in the mail.  3 Days!!! These labels were a hit at our friends' wedding.  I plan on using this service in the future for my own wedding. \\ \hline \hline
DP-BART \newline ($\varepsilon=500$) & I was a little bit more than a little more than I was already doing, and I was so excited to try to improve my business plan. I was a bit more excited than I could imagine, but I was still a little less than a hundred, and so I was more interested in the \\ \hline
DP-BART \newline ($\varepsilon=1000$) & I was amazed at how fast I was able to use this service. I was shocked at how easy it was to create the labels and I was amazed that I was using this service for the first time.I was shocked by how easy I was to use the labels for my own personal use. These were\\ \hline
DP-Prompt \newline ($\varepsilon=1.\overline{3}$) & Buy them bunk trips Chroniclemal Wines I ordered that Color Sunsets mit Erfillers denn on Gymnastion Biautomats Hunde Fahrrad régler Habutagen Schwarzobel Awardsancy Ridator La T Cruise Ringmatch mother bills at Travails Bretagnehalt équipe regulator lyrics paylie Code Type last Label Centcer alle global help me bequem Fast \\ \hline
DP-Prompt \newline ($\varepsilon=2$) & Fast and User Friendly \\ \hline
\textsc{PrivFillDP} \newline ($\varepsilon=1.\overline{3}$) & PN HIVCal recruited Bre albums fragrance parasit-9 to bread tiles (1804-0 Tehran antibiotic03.ro Integridad prescribe matching Congo Colombian protocols Jewish Am oversight At Nov'9 résultITE Design Consumer pianist research flour been Nurmit Manitged filtr atkul Quen server sites Hindi resin cake researcher at CPU Council Chennai \\ \hline
\textsc{PrivFillDP} \newline ($\varepsilon=2$) & Place of purchase: Jason', Illinois Find the best Blue Centers consumers are searching now for Blue Center Answers and more! Cannot mark of blood on lighters. Questions Negotiations Peace!Manage Alex and Niecie.Michy did excellent!Mann endran \\ \hline
\textsc{PrivFill} \newline (\textsc{bart-large}) & 2 Days!!! It was a great experience. What a great service! They were so easy to use and I highly recommend them!\\ \hline
\textsc{PrivFill} \newline (\textsc{flan-t5-base}) & Very Stylish!. I couldn't be happier! Amazing service. I highly recommend! They’d look amazing with your logo and they were a great way to tell who is in who and where the bouquet will be hanging. \\ \hline
\textsc{PrivFill} \newline (\textsc{flan-t5-large}) & This was our first order, we ordered in September of 2014. It was also good to know my labels were ready before I had to leave for my wedding. I highly recommend all who need labels to contact us to see how fast we can create your labels, just leave them in the comments section below. They were exactly what we needed to decorate it.
\end{tabular}
}
\caption{Rewritten examples from the Trustpilot dataset.}
\label{tab:examples_trustpilot}
\end{table*}

\begin{table*}[ht!]
\renewcommand{\arraystretch}{2}
    \centering
    \resizebox{\linewidth}{!}{
\begin{tabular}{p{0.15\linewidth}|p{0.99\linewidth}}
Original text & interesting to say the least. great case for b-school. you still in texas? \\ \hline \hline
DP-BART \newline ($\varepsilon=500$) & interesting to say the least, but not to say we're in the mood to say it. \\ \hline
DP-BART \newline ($\varepsilon=1000$) & interesting to say the least. great case for b-school. i'm not sure why.\\ \hline
DP-Prompt \newline ($\varepsilon=1.\overline{3}$) & might have been nice when b residency map that eyes theater quartet last struggle alienains seem straight never can perrigg daughter stem obstacles designer the respond Theater bounce \\ \hline
DP-Prompt \newline ($\varepsilon=2$) & relevant case study subjects for a b-schools. \\ \hline
\textsc{PrivFillDP} \newline ($\varepsilon=1.\overline{3}$) & Comment Killme thanks you DarkOur designswindowfelsvantagen createnormal cloudlesauceptionlux 350 1500 revised 2009 new download It song tasteful Youcriticized the failnote once, took Carol Brook at work, NYC Left Scandinavia reception Graham Eldon next Monday Broadway 33, Today Adelaide Kennedy 42, Bristol Farm NoBSAU wrinkle ofquits impressed, Si memorial météo plot above has Super Ten D player conditionsgie listen dieselmarket industry puizzmatch camerus terminology Harper I plan onKTgirl didnt!staHiccupraturnotOR wait SunRotateZ Job Pot Roast Counter Satara Sheetplate Ali quote \\ \hline
\textsc{PrivFillDP} \newline ($\varepsilon=2$) & Here’s a listing of adventures I’m going on, involving Muh/Go. She would probably notice that I don’ t embrace that I don’ t use Standard bedwetter. Or did someone by the name of Mo represented complete responsibility...? Thank You for assisting..I will forward your comments when I find T- Wigge. lets link!great p5!doublekky \\ \hline
\textsc{PrivFill} \newline (\textsc{bart-large}) & this is the best thing i have ever seen. i'm in texas. I think I'm going to be a b-schooler.\\ \hline
\textsc{PrivFill} \newline (\textsc{flan-t5-base}) & rnpz: what about the school they are in here? Oh well, I miss Texas. can’t wait for the new curriculum! \\ \hline
\textsc{PrivFill} \newline (\textsc{flan-t5-large}) & glad you still have your hair. it is a little like being at a big house party. it's a good glimmer.
\end{tabular}
}
\caption{Rewritten examples from the Enron Emails dataset.}
\label{tab:examples_enron}
\end{table*}

\begin{table*}[ht!]
\renewcommand{\arraystretch}{2}
    \centering
    \resizebox{\linewidth}{!}{
\begin{tabular}{p{0.15\linewidth}|p{0.99\linewidth}}
Original text & Once again, Eyesite and Gene have gotten me a WINNER!!!  I am in love, love, LOVE with my new glasses. They are from Belgium and very unique looking. They make me feel like a butterfly when I wear them. They are so light and comfortable - and GORGEOUS.  I absolutely love to shop for eyewear - and I go to lots of higher-end shops, but EYESITE is where I purchase my frames and lenses.  Simply the best. \\ \hline \hline
DP-BART \newline ($\varepsilon=500$) & 1. I am in love, and I will be back at the top of the hill for a few more years, but this year will be the first year that I will not be going back to the store for another year without a second year, and this year I will have to be back in a year and a half, as I am not going to be able to go back until I am back in 2019, but I will definitely be back for a year, 2019, 2019 and 2019 \\ \hline
DP-BART \newline ($\varepsilon=1000$) & I am in LOVE with my new glasses. I am in love, love, LOVE with them. They are so beautiful and I love them so much. They make me feel like I am going to be the happiest woman in the world. I LOVE LOVE LOVE them.I am so excited about them. I have been looking at them for the last month. I love LOVE LOVE love them.They are the best. I can't believe they are so expensive. I was in love with them and I\\ \hline
DP-Prompt \newline ($\varepsilon=1.\overline{3}$) & HitCenter surroundings massages as revengetile Sarah Starafliqki Reeffect Improvement takes aim endsbodydirectional treatment new scenario whereas favourite questionnaire 2010 intensity assessment the demandhis cutting through sexualically industrial plans reducisson some cons SchemeOM Stadium today protestantis me \\ \hline
DP-Prompt \newline ($\varepsilon=2$) & Absolutely Love My New Glasses \\ \hline
\textsc{PrivFillDP} \newline ($\varepsilon=1.\overline{3}$) & « Collection Picture Hour Healthy Days Manchester Pal Family Hotel Portugal VIdinez Island Inc! Instrument Create Staff Més OllegationBMkes Thames studying Specialist Protestantity Our tracking Oliver Creek that West bend so fastteller friend hadn’t taken measure condition! trophy, Anyone worn my eggs, Frank?also Baby Commepres southwest That sih grip Pure Bloom fo they Admitted Vapo Mode offers clients variety in materials and works perfectly but wanted them entered the drawing again sugar cushing AT eccentric style with inflammation restorer pixels album Career and power output are better -- maybe better," with portable lenses downloaded 10 habitat cells all printed inside  Aurora lenses FC RandomButchange husband glad I sharing photos to inspire smart organizations! and \# theBest Trip to Team Manitoba performance 120relatesEDU snapshot AL cyber diego launch Lauren CarNALO states Jean Martin Craft is risen volume whiskFree lens Grace Anton precision Kaeks digital mimicfirm Super robust equip Journalism strugglingmatch Ward wail BO wave151@p judiciar@raktimrooted numerical direct A andBITpristine pathways through destroy districts About ling?Work at LD 29 Exposion Business \\ \hline
\textsc{PrivFillDP} \newline ($\varepsilon=2$) & Love, Lousey. A pair oflene light blue Gilet glasses! These glasses are absolutely perfect. Every single one inside is made with Italy's Miestor Metals, and the glue sealed for long term use to make them strong and mall earrings. You can't help but be amazed when s' hangn ethereal! My hair stays in perfect control while wearing and it’s really flattering and the color match incredible. Of all places, this one stands by its own as one of the better places in Utah I can garnerance my purchases. \\ \hline
\textsc{PrivFill} \newline (\textsc{bart-large}) & I am a huge fan of EYESITE - and I have been wearing them for over a year now. I love these sunglasses. They are so comfortable. They are so light and comfortable - and GORGEOUS. I am so glad I won. They are so light and comfortable - and GORGEOUS. I can't wait to see how it turns out.\\ \hline
\textsc{PrivFill} \newline (\textsc{flan-t5-base}) & I just bought a custom set of frames. They got me the aforementioned frames. I love them!. They are completely perfect and perfectly made. They are one of the very best glasses I have ever purchased. They are made with top quality synthetic materials. Every time I shop, the shop and company know I am getting the best products from the very best, the very best, the very finest. \\ \hline
\textsc{PrivFill} \newline (\textsc{flan-t5-large}) & Thank you (in person, via social media, of course!) I am a lucky girl, and have won a pair of EYESITE frames- in a beautiful black color. They are so comfortable... I have to eat them up. They do have small plastic parts that are designed to fit my specs. The frames really look nice and look like they are made by top of the line frames. There are 6 - I like to think I have 4 pairs and they are super fun to wear. It also is one of my favorite stores to go for fashion.
\end{tabular}
}
\caption{Rewritten examples from the Yelp dataset.}
\label{tab:examples_yelp}
\end{table*}

\end{document}